\documentclass[journal]{IEEEtran}

\usepackage{times}
\usepackage{epsfig}
\usepackage{graphicx}
\usepackage{amsmath}
\usepackage{amssymb}
\usepackage{cite}
\usepackage{multirow}
\usepackage{comment}
\usepackage{color}
\usepackage{threeparttable}

\usepackage{color}
\definecolor{citecolor}{RGB}{119,185,0} 
\usepackage[pagebackref=false,breaklinks=true,letterpaper=true,colorlinks,citecolor=citecolor,bookmarks=false]{hyperref}
\usepackage{colortbl}

\usepackage{amsmath}

\def\eg{\emph{e.g.}} 
\def\ie{\emph{i.e.}} 
\def\etal{\emph{et~al.}} 

\newlength\savewidth\newcommand\shline{\noalign{\global\savewidth\arrayrulewidth
  \global\arrayrulewidth 1pt}\hline\noalign{\global\arrayrulewidth\savewidth}}

\usepackage{floatrow} 
\begin{document}

\title{  Self-supervised Point Cloud Representation Learning  via Separating Mixed Shapes}

\author{Chao Sun, Zhedong Zheng, Xiaohan Wang, Mingliang Xu and Yi Yang,~\IEEEmembership{Senior~Member,~IEEE} 
\thanks{Chao Sun, Xiaohan Wang and Yi Yang are with School of Computer Science, Zhejiang University, Zhejiang, China. E-mail: c\_sun@zju.edu.cn, xiaohan.wang@zju.edu.cn, yangyics@zju.edu.cn}
\thanks{Zhedong Zheng is with Sea-NExT joint lab, School of Computing, National
University of Singapore, Singapore 118404. E-mail: zdzheng@nus.edu.sg}
\thanks{Mingliang Xu is with Zhengzhou University, 100 Kexue Ave, Zhongyuan District, Zhengzhou, Henan, China. Email: iexumingliang@zzu.edu.cn}
\thanks{This work is supported by National Key R\&D Program of China (No.2020AAA0108800) and Fundamental Research Funds for the Central Universities (No. 226-2022-00087).}
}

\markboth{Journal of \LaTeX\ Class Files,~Vol.~14, No.~8, August~2015}%
{Shell \MakeLowercase{\textit{et al.}}: Bare Demo of IEEEtran.cls for IEEE Journals}
\maketitle

\begin{abstract}
The manual annotation for large-scale point clouds costs a lot of time and is usually unavailable in harsh real-world scenarios. Inspired by the great success of the pre-training and fine-tuning paradigm in both vision and language tasks, we argue that pre-training is one potential solution for obtaining a scalable model to 3D point cloud downstream tasks as well. In this paper, we,  therefore,  explore a new self-supervised learning method, called Mixing and Disentangling (\textbf{MD}), for 3D point cloud representation learning. 
As the name implies, we mix two input shapes and demand the model learning to separate the inputs from the mixed shape. 
We leverage this reconstruction task as the pretext optimization objective for self-supervised learning. 
There are two primary advantages: 1) Compared to prevailing image datasets, \eg, ImageNet, point cloud datasets are \emph{de facto} small. The mixing process can provide a much larger online training sample pool. 
2) On the other hand, the disentangling process motivates the model to mine the geometric prior knowledge, \eg, key points. 
To verify the effectiveness of the proposed pretext task, we build one baseline network, which is composed of one encoder and one decoder. During pre-training, we mix two original shapes and obtain the geometry-aware embedding from the encoder, then an instance-adaptive decoder is applied to recover the original shapes from the embedding. 
Albeit simple, the pre-trained encoder can capture the key points of an unseen point cloud and surpasses the encoder trained from scratch on downstream tasks. 
The proposed method has improved the empirical performance on both ModelNet-40 and ShapeNet-Part datasets in terms of point cloud classification and segmentation tasks. We further conduct ablation studies to explore the effect of each component and verify the generalization of our proposed strategy by harnessing different backbones.

\end{abstract}



\begin{figure}[t]
\begin{center}
\includegraphics[width=0.95\linewidth]{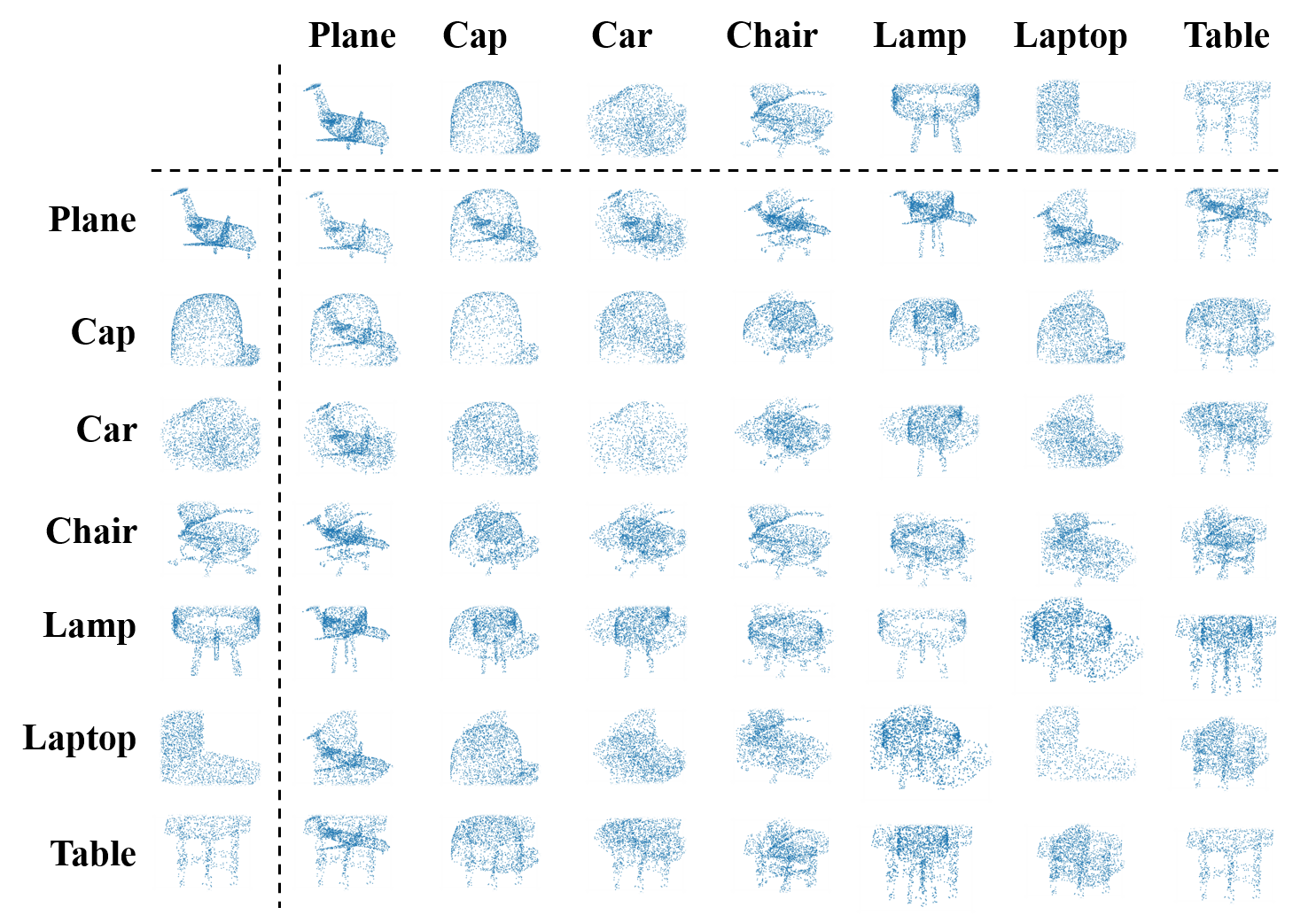}
\end{center}
\vspace{-.15in}
   \caption{ Visualized results of mixed point clouds. We select seven types of point clouds from ShapeNet-Part~\cite{yi2016scalable}. Each row and column corresponds to the original point clouds and the intersection corresponds to the mixed point cloud. $M$ denotes the number of samples in the training set. \textbf{In theory, we can generate $O(M \times M)$ different point clouds by sampling various cloud pairs, resulting in a much larger online generated training sample pool.} }
\label{fig:Fig.1}
\end{figure}

\section{Introduction}
\IEEEPARstart{P}{oint} clouds, as one perception of the 3D world, have wide applications, \eg, autonomous driving~\cite{cui2021deep, li2020deep, tampuu2020survey} and virtual reality~\cite{guo2020deep}. With recent developments of deep learning technology, deep learning-based approaches~\cite{wan2021rgb, chen2020boost, wang2018pixel2mesh, hu2021learning} on point cloud processing gradually surpass traditional statistical processing methods~\cite{guo2020deep, huang2021comprehensive}. 
However, these deeply-learned models are data-hungry. Although the point cloud data can be collected by laser sensors and other equipment, the point-wise point cloud annotation still costs a lot of human resources and unaffordable expenses~\cite{wu20153d, yi2016scalable, armeni2017joint, dai2017scannet}. In this work, we argue that pre-training is one potential way to relieve data limitation. We are inspired by successes in image recognition, where the pre-training model on ImageNet can efficiently adapt to various computer vision tasks, including image segmentation~\cite{2015U, long2015fully, 2016Pyramid, 2018Encoder} and image retrieval~\cite{2016Deep, 2017Large, 2017Unlabeled, zhang2019deep,zheng2020vehiclenet}. We also note that pre-training on point cloud is still under-explored. 
To fill this gap, in this work, we resort to model pre-training via self-supervised learning to reduce the demand for annotated data. 

\begin{figure}[t]
\begin{center}
\includegraphics[width=0.95\linewidth]{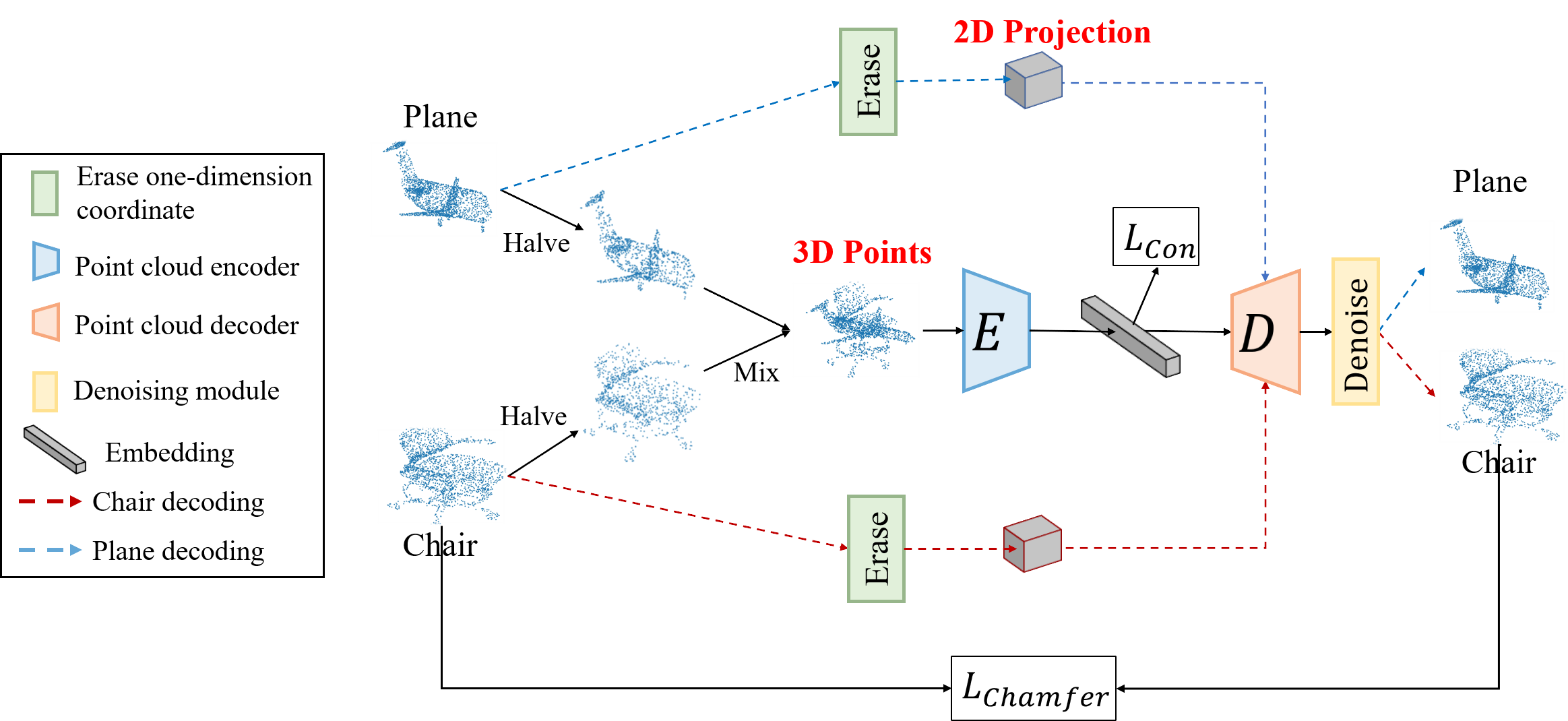}
\end{center}
\vspace{-.15in}
   \caption{A schematic overview of our method. The two original point clouds are mixed and input into encoder $E$ to obtain the embedding vector of the mixed point cloud. A coordinate extracting operation, $Erase$, is used to extract partial information from the original point cloud for instance disentanglement. Then the embedding vector and instance information are input to the decoder $D$ and the denoising module to generate the original point cloud, as shown in the \textcolor{red}{red} line and the \textcolor{blue}{blue} line, respectively. 
   The reconstruction loss, \ie, $L_{Chamfer}$, between the generated point cloud and the original point cloud is used as the self-supervision during training.}
\label{fig:overview}
\end{figure}

Most existing works~\cite{sauder2019self, xie2020pointcontrast} focus on designing pretext tasks by exploring the spatial characteristics of the single point cloud, which does not solve the problem of data limitation in current open-source datasets~\cite{wu20153d, yi2016scalable}. 
To address this limitation, we introduce a simple solution Mixing to combine two point clouds as a new mixed point cloud.
Benefiting from the batch training of deeply-learned models, each point cloud can be mixed with a large number of point clouds during the whole training process to enlarge the training data pool. The proposed approach is memory and resource-efficient, which can directly process point clouds in a single pass.
Compared with texture, which is important in 2D images, shape information is critical for point cloud representation. The human can easily figure out the two 3D models, \ie, chair and airplane, in mixed point clouds, according to the prior knowledge of shape (see Fig.~\ref{fig:Fig.1}). 
Inspired by the work on 2D clothes changing~\cite{zheng2019joint}, we propose a novel self-supervised task, Mixing and Disentangling (\textbf{MD}) for point clouds. 
As the name implies, our intuition underpinning the proposed \textbf{MD} is encouraging the network to mine geometric knowledge, \ie, shape-aware features, and such knowledge can be easily transferred to various tasks. 
As shown in Fig.~\ref{fig:overview}, the pretext task of self-supervised learning is designed to separate original point clouds from the mixed one.
Given two input point clouds A of $N$ points and B of $N$ points, the mixing progress outputs mixed point cloud C of $N$ points sampled from A and B. In particular, we adopt a random sampling strategy in each input point cloud and select $\frac{N}{2}$ points in A and B respectively. After the sampling stage, we concatenate 2 sampled $\frac{N}{2}$ points together as a new point cloud C with $N$ points, and we also disrupt the indices of these points again to prevent over-fitting the point order.
The disentangling process demands the model to mine the key points of both two original point clouds from the mixed point cloud.
The decoder output 2 tensors shape of $N\times3$, and each reconstructs one input point cloud. In particular, the decoder disentangles a specified point cloud based on its conditional coordinates shown in Fig.~\ref{fig:2coords} as the decoder input. For instance, two generated point clouds shown in Fig.~\ref{fig:overview} demands two decoding process (the model should forward twice while the sum loss of two results backward once), one for the plane, and another for the chair.
Briefly, our contributions are as follows: 
\begin{itemize}
\item Different from most existing pre-training works on image recognition, there do not exist large-scale datasets like ImageNet~\cite{ILSVRC15} for 3D point cloud pre-training. To address this problem, we leverage the mixing process to generate large-scale mixed data for 3D point cloud training;

\item Inspired by the human ability to figure out two shapes from one mixed object, we propose a new self-supervised learning method on the point cloud, called Mixing and Disentangling (\textbf{MD}) to learn the geometric prior knowledge without the requisite of annotations;

\item As one minor contribution, we implement one basic pipeline to verify the effectiveness of the proposed self-supervised learning strategy. It contains one encoder with the learnable aggregation function and one instance-adaptive decoder, to learn from the mixed point cloud and conduct the disentanglement. 

\item Albeit simple, experimental results on two benchmarks, \ie, ModelNet-40~\cite{wu20153d} and ShapeNet-Part~\cite{yi2016scalable}, show that the self-supervised learning model can effectively and efficiently improve the accuracy of classification and segmentation tasks by the pre-training and fine-tuning paradigm. Self-supervised learning on the point cloud also can reduce the network dependence on labeled data. 

\end{itemize}

The rest is organized as follows. We introduce existing works in Section~\ref{sec:related}. Section~\ref{sec:method} describes the proposed method with details. Quantitative and qualitative experiments verify the effectiveness of \textbf{MD} pre-training in Section~\ref{sec:exp}, followed by the conclusion in Section~\ref{sec:conclusion}. 

\section{Related Work}\label{sec:related}

\subsection{3D Point Cloud Processing}
In recent years, deep learning-based approaches~\cite{wang20183d, yi2017large, meng2021towards, chen2020pointmixup} facilitate the development of point cloud processing. Due to the irregular and disordered characteristics of point clouds, the traditional neural networks in the 2D field, \ie, convolutional neural network (CNN), cannot be directly applied to point clouds. Therefore, researchers resort to various methods as follows.  \textbf{(1).} Some methods are voxel-based. The voxel-based methods~\cite{maturana2015voxnet, wu20153d} voxelize the point cloud to obtain a cube with a grid structure, which can be directly processed by 3DCNN. However, the accuracy of this method is limited by the resolution during voxelization. \textbf{(2).} Some methods are projection-based. The projection-based methods~\cite{su2015multi, li2020auto, zhang2020polarnet} project the point cloud on multiple planes. The existing 2D CNN is used to extract the projected point cloud features on different planes. After fusing multiple features, the descriptor of the point cloud is obtained, which can be used in downstream tasks.  
\textbf{(3).} Some methods are point-based. PointNet~\cite{qi2017pointnet} proposed a neural network structure that directly processes the raw point cloud data. PointNet++~\cite{qi2017pointnet++} further considers both the global feature and the local feature. Point-based approaches can be divided into the graph-based method and the convolution-based method. The graphed-based method treats the point set as a graph~\cite{kipf2016semi, hu2020randla, wang2019dynamic}. The convolution-based method~\cite{thomas2019kpconv, xu2021paconv} introduces the concept of the convolution kernel in 2D CNN to the point cloud. Taking one step further, recent methods~\cite{guo2021pct,zhao2021point} also explore attention and transformer structures, yielding competitive performance. 
    
  \subsection{Self-Supervised Learning}
    
Self-supervised learning is a machine learning paradigm that uses the structural information of the data itself to generate labels required for training. 
\textbf{(1).} Some methods are based on the 
transformation invariance. In particular, image rotations~\cite{gidaris2018unsupervised}, image jigsaw puzzle~\cite{noroozi2016unsupervised}, random erasing~\cite{zhong2020random} , adaptive exploration ~\cite{ding2020adaptive} 
have been shown to be helpful. 
\textbf{(2).} Some methods are based on the context of data. This type of method learns the feature by image generation, \eg, image colorizing~\cite{zhang2016colorful}, image inpainting~\cite{2016Context} and super-resolution methods~\cite{2016Photo}. \cite{zhu2021temporal} adopts a context reconstruction loss for self-supervised temporal modeling.
Generative Adversarial Networks~(GANs) are used for image generation~\cite{2014Generative, zhu2017unpaired,zheng2019joint}. 
\textbf{(3).} Other methods are based on contrastive learning. The type of method mines the structure information from data by designing loss functions. These methods define positive and negative samples firstly and then conduct metric learning by narrowing the distance between positive pairs and widening the distance between negative pairs. For instance, CPC~\cite{oord2018representation} introduces anti-noise estimation, while InfoNCE~\cite{oord2018representation} proposes a loss function based on mutual information. 
Following the spirits of these works, researchers~\cite{chen2020simple,he2020momentum,han2020Self,grill2020Bootstrap,zbontar2021barlow} further design various model architectures and loss functions.
Even worse than 2D image datasets, the annotation of 3D point clouds is usually unavailable due to the annotation time costs and human expenses. In recent years, there are several self-supervised learning methods~\cite{zhang2019unsupervised, han2019multi, rao2020global, wang2020unsupervised1}, which explore the point cloud pre-training. For instance, Sauder~\etal~\cite{sauder2019self} propose space reconstruction, PointContrast~\cite{xie2020pointcontrast} is based on multi-view contrastive learning, and DepthContrast~\cite{zhang_depth_contrast} leverages contrastive learning which can handle different input data formats. 
However, these methods usually require a relatively large dataset, such as Scannet~\cite{dai2017scannet}, and do not solve the data limitation in real-world tasks. To fill this gap, we adopt a simple point cloud mixing strategy to enlarge the small dataset. Some works have already applied the mixing progress to supervised learning as a data augmentation strategy. For example, PointMixup~\cite{chen2020pointmixup} proposes the shortest path interpolation with EMD~(Earth Move Distance), while PointCutMix~\cite{zhang2021pointcutmix} searches the one-to-one correspondence by EMD. 
Besides, RSMix~\cite{lee2021regularization} proposes to replace some points by extracting subsets from another point cloud. 
Differently, our method focuses on self-supervised training of point clouds with the disentangling task while these works only apply the mixing progress as a data augmentation in supervised learning. 
From another point of view, our method is complementary to the existing work.
We can apply our method to conduct self-supervised pre-training on the unlabeled dataset and fine-tune the model on the downstream task with the above data augmentation strategies. 
\textbf{The main difference between the proposed method and existing works are: 1). We generate more ``within-distribution’’ training point clouds via Mixing, which have the same mean and variance with the original data. This process largely increases the training pool and lets the model ``see'' more inlier variants during training. 2). The proposed pretext task is more challenging. Separating original objects from mixed point clouds is more difficult than extracting key points of a single object. 3). We 
show that the learned models are competitive in various downstream tasks and settings.}

\begin{figure}[t]
\begin{center}
\includegraphics[width=0.95\linewidth]{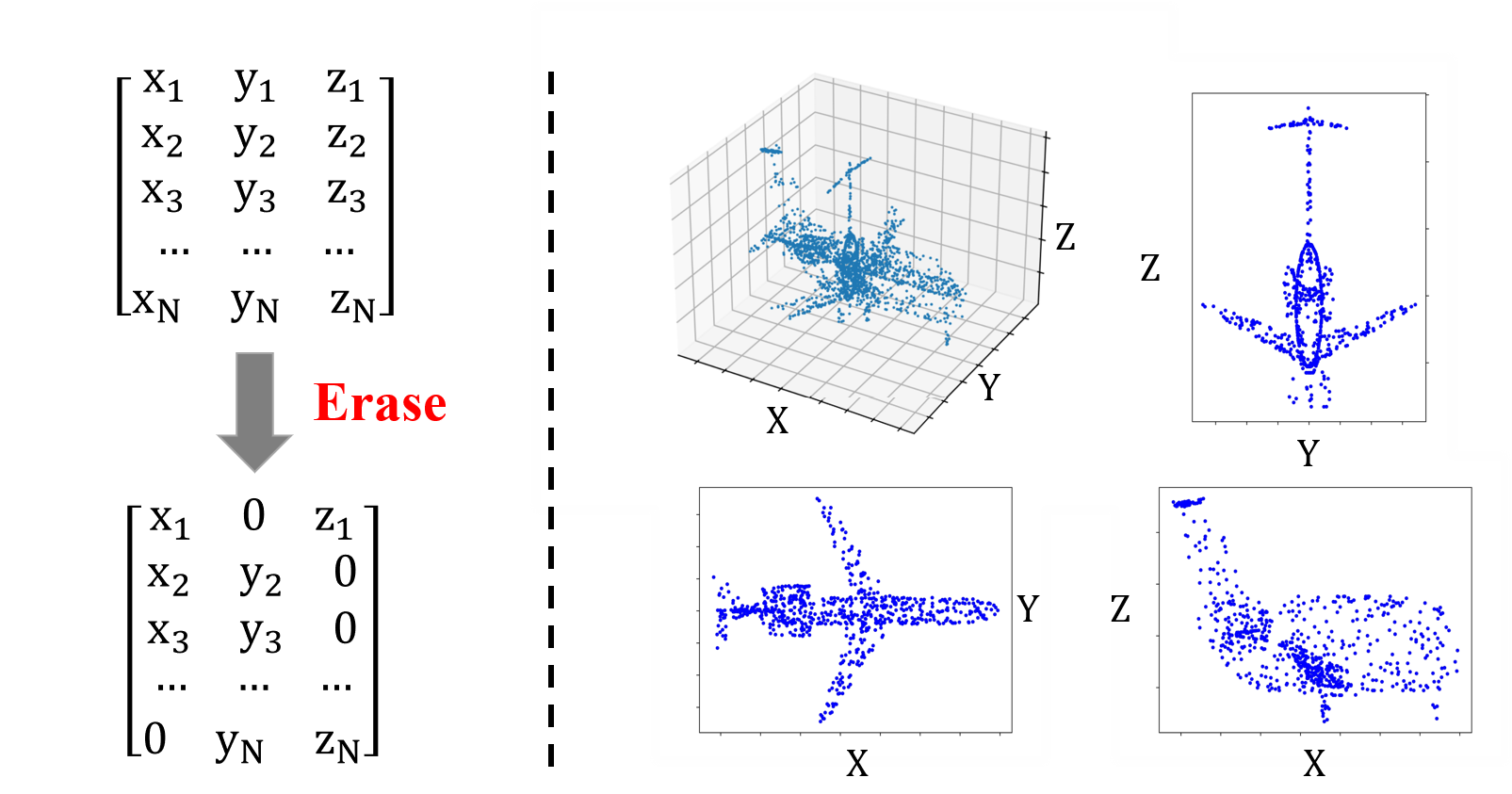}
\end{center}
\vspace{-.15in}
   \caption{The left part represents a point cloud of $N$ points, and each point has 3D coordinates. After erasing, the random one-dimensional coordinate of each point is set to zero which intends to make the pre-training process more challenging. The right part is the visualization of an erased point cloud of the plane and its three-view drawing after erasing. The erasing process equals randomly set about $\frac{1}{3}N$ points to the corresponding projection surface.}
\label{fig:2coords}
\end{figure}

\begin{figure*}[t]
\begin{center}
\includegraphics[width=0.95\linewidth]{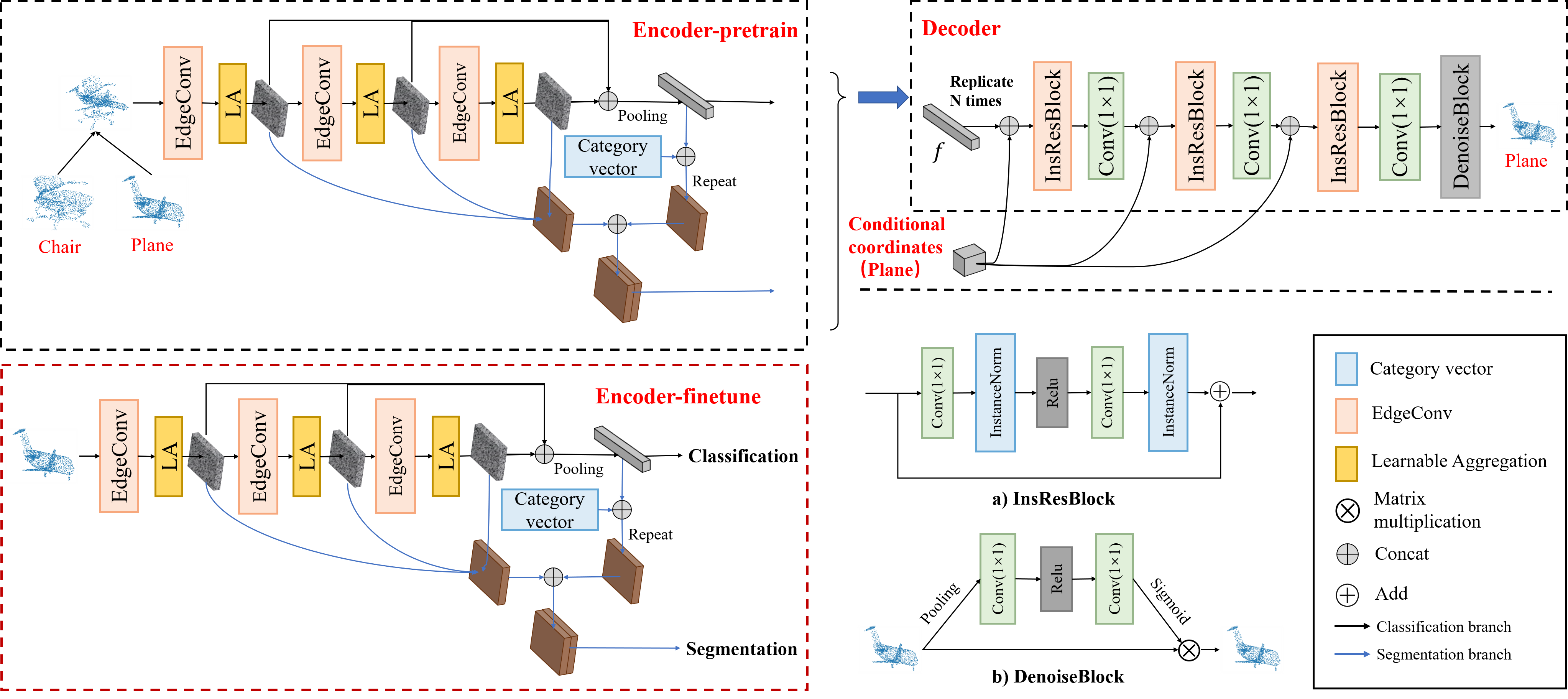}
\end{center}
\vspace{-.15in}
   \caption{\textbf{Pipeline.} The model is mainly composed of one encoder (left) and one decoder (right). \textbf{Encoder: }The encoder takes $N$ points as input and aggregates the special features of each point and its corresponding points in the neighborhood at an EdgeConv layer. The classification network (top branch) contains three EgdeConv layers and three learnable aggregation (LA) layers. The output of the encoder is the embedding vector $f$ of the input point cloud. In the segmentation network, to concatenate with the point-wise feature from the shallow layers, the corresponding embedding vector $f$ is repeated $N$ times for point segmentation, where $N$ is the number of points. 
   The feature maps from shallow layers are concatenated with the embedding vector to obtain the multiple-scale feature map as the output of the encoder. Different from the classification branch, the category vector representing the type of point cloud is fused into the embedding vector for the part segmentation. 
   \textbf{Decoder:} The decoder is composed of Instance-aware Residual Block (InsResBlock), $1\times1$ Convolution Layers and one DenoiseBlock to refine the final reconstructed point cloud. The structure of InsResBlock and DenoiseBlock is shown in a) and b). The random two-dimensional coordinates of the original point cloud are used as conditional information to be fused with the backbone features in three different layers.  }
\label{fig:encoder-decoder}
\end{figure*}

\section{Methodology}
\label{sec:method}
In this section, we illustrate the pipeline and the basic model, which simply contains two modules, \ie,  encoder, and decoder. Then we explain the training strategy of Mixing and Disentangling, followed by the discussion on components.

\subsection{Overview}
\textbf{Our model is to recover two 3D input point clouds according to the corresponding 2D projections from the mixed point cloud. This challenging task motivates the model to learn shape-related geometric knowledge, which benefits downstream tasks.}
As shown in Fig.~\ref{fig:overview}, the structure is mainly composed of one encoder and one decoder. The encoder is to extract the embedding vector of the mixed point cloud. Given the output features from the encoder, the decoder is used to restore the original point clouds based on the embedding vector and the conditional input from partial coordinates of the original point cloud ~(see Fig.~\ref{fig:2coords}).
After pre-training, the encoder can be utilized to extract discriminative features for subsequent tasks, \eg, classification, and segmentation. As shown in Fig.~\ref{fig:encoder-decoder}, the black arrows denote the fine-tuning workflow for recognition, while the blue arrows are the segmentation workflow. 
We denote the input point cloud as a set of points $S=\{s_1, s_2, ... ,s_n\}$, $s_i \in \mathbb{R}^3$ and each point has 3D coordinates. However, the point cloud does not have a regular spatial structure as 2D images, we can not apply the convolutional neural network~(CNN) directly. 
Therefore, we apply K-Nearest Neighbor~(KNN) algorithm to construct the graph structure $\mathcal{G}$ in the feature space. The KNN algorithm based on the feature inputs can efficiently and effectively find two points with the most similar semantics, \ie, the points on the two legs of the chair with similar semantics. The graph can be expressed as $\mathcal{G}=(\mathcal{V},\mathcal{E})$, where $\mathcal{V}$ represents the set of vertices and $\mathcal{E}\subseteq \mathcal{V} \times \mathcal{V}$ represents the edge set. 

\subsection{Encoder}
As shown in Fig.~\ref{fig:encoder-decoder}, the backbone of the encoder is composed of three EdgeConv layers and three learnable aggregation (LA) layers. Specifically, we deploy EdgeConv layers to leverage the graph information. Given the input of $N\times H_{in}$ and the corresponding graph $\mathcal{G}$, the EdgeConv layer performs feature transformation on the $N$ points and outputs the updated point features of $N\times H_{out}$. The input features can be represented as $X=\{x_1, x_2,...,x_n\}$, $x_i \in \mathbb{R}^{H_{in}}$ where $x_i$ represents the feature of point $i$. The relation feature $r_i^k \in \mathbb{R}^{N \times H_{out}}$ can be formulated via an EdgeConv layer as:
{\setlength\abovedisplayskip{0.15cm}
     \setlength\belowdisplayskip{0.15cm}
\begin{equation}
r_i^k=MLP(concat(x_i, x_i-x_k)),k:(i,k)~\in \mathcal{E}, k \neq i,
\end{equation}}
explicitly encodes the relationship between $x_i$ and $x_k$, where $x_k$ is one neighbor of $x_i$ and $MLP$ notes a linear function with ReLU~\cite{nair2010rectified}.
To consider all the neighbor of $x_i$, we apply Learnable Aggregation (LA) to the local neighbor relation $r^k_i$. 
We notice that the current widely-used aggregation methods applied in the point cloud include both max pooling and average pooling. It is usually challenging to decide which pooling function should be used. As one minor contribution, we involve Learnable Aggregation (LA), which is a learnable pooling layer, and let the model learn the adaptive weight:
    {\setlength\abovedisplayskip{0.15cm}
     \setlength\belowdisplayskip{0.15cm}
    \begin{equation}
        LA(x_i) = \alpha \times \mathop{max}_{(i,k)~\in \mathcal{E}, k \neq i}(r_i^k) + (1-\alpha)\times \mathop{avg}_{(i,k)~\in \mathcal{E}, k \neq i}(r_i^k),
        \label{lmmp}
    \end{equation}
    }
where the interpolation ratio $\alpha$ is a learnable parameter of the network. 
Finally, the encoder backbone outputs the point cloud feature map by combining intermediate features from multiple layers as~\cite{yang2021multiple} to enhance the representation capability.
For the classification downstream tasks, we add the max pooling layer to compress the feature map to the vector, which is sent to the decoder. When fine-tuning the downstream classification, we simply add one linear classifier to this learned vector. 
As for the segmentation downstream task, which is a dense point-level task, the vectorization compromises the spatial representation.  Therefore, we concatenate the intermediate feature maps to keep the same structure for fine-tuning. We also add the extra category vector to specify part predictions, because different objects contain different semantic parts. For instance, the plane contains the aircraft nose, plane body, plane tail, and wings while the chair has legs, the seat, and the back. When fine-tuning, similarly, we only need to add one linear classifier to the learned feature map. 

\textbf{Discussion.} By using the proposed mixing mechanism, we can generate more ``within-distribution’’ point cloud samples, which largely enriches training samples. The previous methods~\cite{sauder2019self} use ShapeNet-Part for pre-training and it can only generate $O(M)$ training samples, where $M$ is the number of point clouds. In theory, our method can generate $O(M\times M)$ different point clouds by sampling various cloud pairs, resulting in a much larger online generated training sample pool. The encoder can mine more prior knowledge from data, which improves downstream tasks. Besides, to extract the features more efficiently, we propose a new feature aggregation method LA in the point cloud. Most existing works ~\cite{qi2017pointnet++, wang2019dynamic} apply max pooling to aggregate the neighboring features, which leads to much information loss. LA leverages a learnable parameter to dynamically adjust the aggregation method, which can effectively retain the local structure. More ablation studies on LA are provided in the experiment.

\subsection{Decoder}

We propose an instance-adaptive decoder, which can restore the input point cloud from the embedding vector according to the conditional coordinates adaptively. The decoder treats the embedding vector as the structure representation of the mixed point cloud and aims to disentangle key points from different objects. In our network, the decoder is utilized as a selective filter, which gradually enhances the feature of points belonging to one object and weakens the feature of the rest points of another object. After several transformations, the feature mainly contains the key points of one object, which can be transformed to the recovered point cloud by the final linear layers. As shown in Fig.~\ref{fig:encoder-decoder}, the decoder is composed of Instance-aware Residual Block~(InsResBlock)~\cite{huang2017arbitrary}, $1\times1$ Convolution Layers and one denoising module \ie, DenoiseBlock. There are two inputs to the decoder: the mixed point cloud embedding vector $f$, the conditional coordinates $C_{coord}$ of size $N\times3$.
To recover the original point cloud from the mixed feature, the 2D projection $C_{coord}$ of each point cloud is given during training.
We use the random two-dimensional coordinates of the original point cloud as partial information. As shown in Fig.~\ref{fig:2coords}, we randomly erase ~\cite{zhong2020random} one of the three coordinates of each point cloud. For each point in a point cloud, the erased dimensions are randomly generated instead of taking a certain dimension. Since the direct erasing of a certain dimension will cause the point cloud to change from $N\times3$ to $N\times2$, the neural network cannot distinguish the specific dimensions of the remaining two coordinates. For instance, the model can not foreknow the two coordinates are (X,Y) or (X,Z) or (Y,Z).
We choose to set the erasing coordinates to zero instead of directly deleting the dimension. Instance-aware Residual Block~(InsResBlock) is a basic module used to disentangle the mixed feature from $f$ according to the conditional coordinates $C_{coord}$. The structure of InsResBlock is shown in Fig.~\ref{fig:encoder-decoder} (a), which is composed of two convolution layers and instance normalization~\cite{ulyanov2016instance}.
A pair of InsResBlock and $1\times1$ convolution layer is regarded as a basic decoding unit. 
Our decoder $D$ is built with three sequential basic decoding units, which are connected sequentially to reduce the feature dimension layer by layer and finally output the generated point cloud of $3$ coordinate channels. For each basic encoding unit, we concatenate the conditional coordinates and the feature map from the previous unit to specify the reconstruction target. 
To refine the generated point cloud, we further introduce a denoising module as the last unit of decoder.  
As shown in Fig.~\ref{fig:encoder-decoder} (b), the denoising module is implemented via a self-attention manner. 
The module leverages the context information between neighbor points to assign different weights for adjacent points and noisy points, which can ensure each point will be optimized with the global information. 
Given the noisy point cloud $\widetilde{s}$, the denoising module outputs  the denoising point cloud $\hat{s} = Denoise(\widetilde{s}), \widetilde{s} = D(f, C_{coord})$,
where $D$ represents the decoder.
In particular, given the tensor $\widetilde{s}$ of $N\times3$, we perform average pooling in the dimension of $N$ to obtain an initial weight of $N\times1$ for every point. 
We apply two convolution layers to aggregate the global information from all points and normalize the weight of each key point via the sigmoid function. Then the weight is multiplied by the original point cloud to obtain the denoised point cloud. The multiplication allows every point to refer to the neighbor points and ignore outlier points with low scores. 
    
\textbf{Discussion.} \textbf{The choice of conditional disentanglement.} The choice of $C_{coord}$ has a large impact on the reconstruction results and downstream tasks. If the decoder can only obtain little original point cloud information from $C_{coord}$, the reconstruction effect of disentanglement will be poor. On the contrary, if the decoder obtains too much information from the original point cloud, the decoder will restore the original point cloud based on the conditional information directly. Then the encoder cannot be well trained to fully mine the geometric information. We think this choice is still open and in this paper, we observe that using random two-dimensional coordinates is a balanced choice, which guarantees the completeness of the disentanglement and the challenge of the pre-training task.

\subsection{Optimization}
As the target of the proposed Mixing and Disentangling task (MD), we supervise the model training via reconstruction.
Besides, we add the intermediate embedding loss for self-supervision. There are many candidates for embedding loss. In this work, without loss of generality, we select contrastive loss to verify the compatibility of the proposed method instead of pursuing the best loss. Next, we describe the two objectives. 

\textbf{Reconstruction loss.}  
We apply Chamfer distance to measure the distance between the original point clouds $s$ and the reconstructed one $\hat{s}$. As Eq.~\ref{chamfer sistance} shows, the distance is a symmetric function. Since we could not know the exact matching between two point sets, the Chamfer distance accumulates one sub-optimal but effective distance between the closest points in both the original and reconstructed point cloud. The reconstruction loss can be formulated as:
    {\setlength\abovedisplayskip{0.15cm}
     \setlength\belowdisplayskip{0.15cm}
    \begin{equation}
        \begin{split}
            L_{Chamfer} = & \frac{1}{|\hat{s}|} \sum_{\hat{p}\in \hat{s}} \mathop{\min}_{p\in s} {||p-\hat{p}||}_2 + \frac{1}{|s|} \sum_{p\in s} \mathop{\min}_{\hat{p}\in \hat{s}} {||\hat{p} - p||}_2,
        \end{split}
        \label{chamfer sistance}
    \end{equation}
    }
where $p$ denotes the point position in $s$, and $\hat{p}$ denote the points in $\hat{s}$. $||\cdot||_2$ denotes the L2 distance between two points and $|\cdot|$ denotes the number of points. 
Specifically, we apply the Chamfer distance as the reconstruction loss on $s$ and $\hat{s}$. 


\textbf{Embedding loss.} Inspired by contrastive loss~\cite{2006Dimensionality}, we introduce it to our total loss function, which intends to widen the distance between different categories.
In particular, since we apply a mix strategy, we view every sample as one single category. 
Following the existing practise, \eg, Instance loss~\cite{zheng2017dual} and MoCo~\cite{he2020momentum}, we encourage that these point clouds should have different embeddings and deploy metric as an optimization objective.
Given a mini-batch of \textit{B} different mixed point clouds, we construct $B\times B$ pairs of samples by calculating the distance $G_{ij}$ between every pair ($i$ and $j$). The basic contrastive loss can be formulated as:
{
\small
\begin{equation}
        \begin{aligned}
            L_{Con} =  \frac{1}{B^2} \sum_{i=1}^{B} \sum_{j=1}^{B} [y{G_{ij}}^2 + (1-y)max(0,{ 1-G_{ij})}^2],
        \end{aligned}
        \label{embedding vector}
\end{equation}
}
We can transfer the distance into the similarity format. Considering $Q_{ij}=1-G_{ij}$, Eq.~\ref{embedding vector} can be rewritten as: 
{
     \small
    \begin{equation}
        \begin{aligned}
            L_{Con} = \frac{1}{B^2} \sum_{i=1}^{B} \sum_{j=1}^{B} [y{(1-Q_{ij})}^2 + (1-y)
             {max(0,Q_{ij})}^2] ,
        \end{aligned}
        \label{Contrastive2}
    \end{equation}
    }
\noindent where $y=1$ only if $i==j$, otherwise $y=0$. In practise, we adopt the cosine similarity ${Q_{ij}=\frac{f_i \cdot f_j}{\left\|f_i\right\| \left\|f_j\right\|}}$. 
Following the previous self-supervised work~\cite{he2020momentum}, we ignore the rare case that $i$ and $j$ belong to the cloud of the same mixed category.
Moreover, we transfer Eq.~\ref{Contrastive2} to the matric format for better efficiency, and adopt L1 loss, which is relatively stable. 
Considering $Q_{ij} \in [-1,1]$, we optimize the distance between $\frac{Q+1}{2}$ and an identity matrix $I$, which can be formulated as: 
    {\setlength\abovedisplayskip{0.15cm}
     \setlength\belowdisplayskip{0.15cm}
    \begin{equation}
        L_{Con} = |\frac{Q+1}{2}-I|.
        \label{Contrastive loss}
    \end{equation}
    }
Finally, to optimize parameters in both encoder and decoder, we deploy the total loss as follows:
{\setlength\abovedisplayskip{0.15cm}
     \setlength\belowdisplayskip{0.15cm}
    \begin{equation}
        L_{total} = L_{Chamfer} + \lambda L_{Con},  
        \label{total loss}
    \end{equation}
}
where $\lambda$ is the weight to control Contrastive loss. Actually, Contrastive loss is an optional choice. In the ablation studies, we also study the effect of Contrastive loss by setting $\lambda=0$. 

\begin{table}[t]
\begin{center}
\vspace{-0.05in}
\footnotesize
\setlength{\tabcolsep}{0.5mm} {
\begin{tabular}{l|c|c|c|c}
\shline
\multirow{3}{*}{Methods} & 
\multirow{3}{*}{Publication} &
\multicolumn{2}{c|}{ModelNet-40} & \multicolumn{1}{c}{ShapeNet-Part} \\
& & \multicolumn{2}{c|}{\emph{Classification}} & \multicolumn{1}{c}{\emph{Segmentation}} \\
\cline{3-5}
& & OA~(\%) & mA~(\%) & mIoU~(\%) \\

\shline
3DShapeNets~\cite{wu20153d}& {\color{black} CVPR'15} & 84.7 &  77.3 & -  \\
VoxNet~\cite{maturana2015voxnet}& {\color{black}IROS'15}& 85.9  &  83.0 & -  \\
PointNet~\cite{qi2017pointnet}& {\color{black}CVPR'17} & 89.2  & 86.0 & 83.7  \\
PointNet++~\cite{qi2017pointnet++} & {\color{black}NeurIPS'17} & 91.9 & -  & 85.1 \\
SpecGCN~\cite{wang2018local}& {\color{black}ECCV'18} & 91.5 & - & -  \\
PCNN by Ext~~\cite{atzmon2018point}&{\color{black} SIGGRAPH'18} & 92.2 & - & 85.1  \\
DGCNN~\cite{wang2019dynamic}& {\color{black}TOG'19} & 92.9 & 90.2 & 85.1  \\
{\color{black}Point Trans.~\cite{zhao2021point}}& {\color{black}ICCV'21} & {\color{black}93.7} & {\color{black}90.6} & {\color{black}86.6}  \\
{\color{black}PAConv~\cite{xu2021paconv}}& {\color{black}CVPR'21} & {\color{black}93.9} & - & {\color{black}86.1}  \\ 
{\color{black}CurveNet~\cite{xiang2021walk}}& {\color{black}ICCV'21} & {\color{black}94.2} & {\color{black}-} & {\color{black}86.8}  \\
 
\hline
Ours~(from scratch)& -  & 92.74 & 89.88 & 85.27 \\
Ours~(pre-training)& - & \textbf{93.39} & \textbf{90.26} & \textbf{85.50}  \\

\shline
\end{tabular}
}
\end{center}
\caption{Results for our baseline and baseline + pre-training. 
We carry out two experiments of classification and segmentation on ModelNet-40 and Shapenet-Part respectively and compare the accuracy with several competitive approaches. 
The baseline model with pre-training achieves the highest accuracy.} 
\footnotesize{
OA: Overall Accuracy; ~
mA: Mean Class Accuracy; ~
mIoU: mean IoU}
\label{table:pre-train}
\end{table}

\section{Experiment} \label{sec:exp}

\subsection{Implementation Details}
\textbf{Datasets.}
(1) ModelNet-40~\cite{wu20153d} is sampled from the mesh surfaces of CAD models, containing 40 categories, 12,311 models. 9,843 models are used for training and 2468 are reversed for testing. Each point cloud contains 2048 points and the coordinates of all points are normalized into the unit sphere. Following existing works~\cite{wang2019dynamic, zheng2020parameter}, we sample 1024 points from each object and augment the data by randomly scaling objects and perturbing point locations. (2) ShapeNet-Part~\cite{2015shapenet} is sampled on the CAD models, having a total of 16,881 point clouds, of which 12,137 point clouds are used for training, 1870 point clouds are used for verifying, and 2874 point clouds are used for testing. Each point cloud is composed of 2048 points, and the coordinates of all points are normalized into the unit sphere. We sample 1024 points from each object. Each point has 4 attributes including the 3D coordinates of each point
and the category label of each point. (3) S3DIS~\cite{armeni20163d} is a real-scan dataset composed of six large scale indoor areas with 271 rooms. Each room is split with $1m \times 1m$ area into blocks. We sample 4096 points from each block. Each point has 9 attributes including 3D coordinates, RGB channels and normalized 3D coordinates of each point. 

\begin{figure}[t]
\begin{center}
\includegraphics[width=0.95\linewidth]{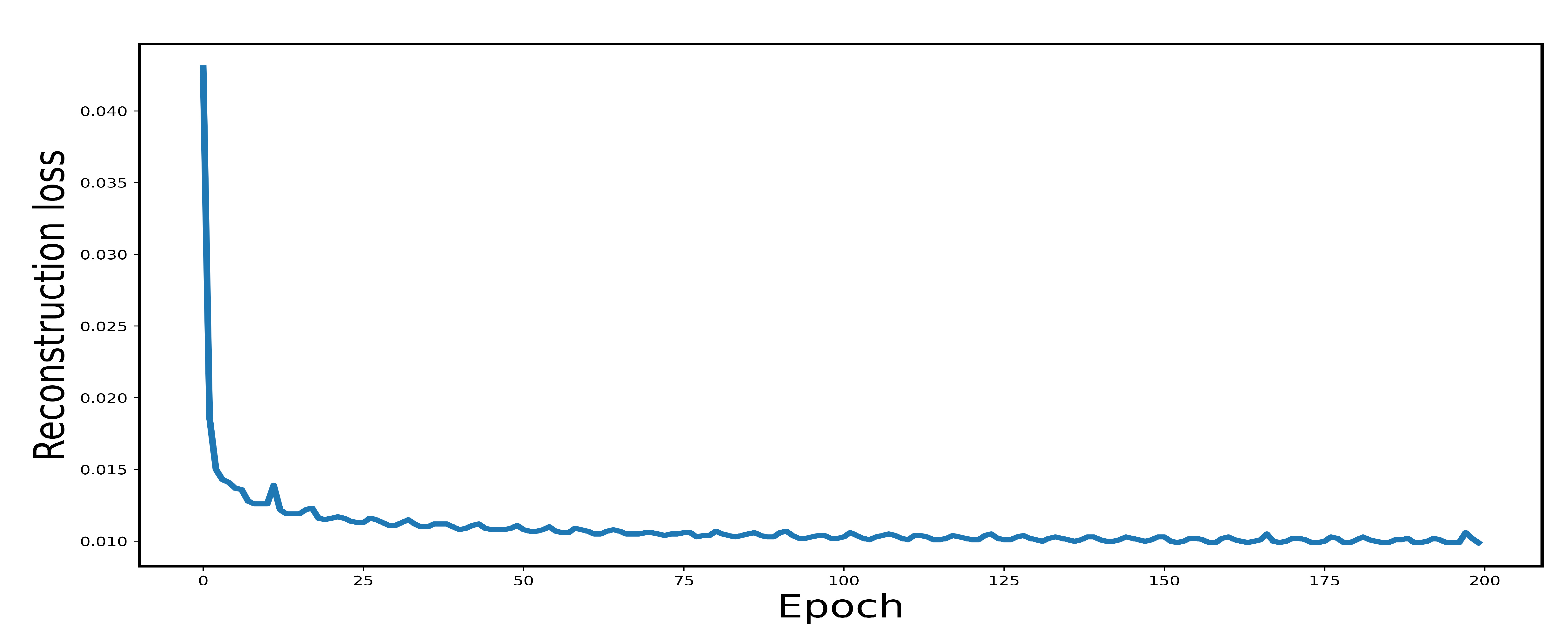}
\end{center}
\vspace{-.15in}
   \caption{The loss curve on ModelNet-40 test set during pre-training. We observe that the proposed method can converge smoothly.}
\label{fig:loss_curve}
\end{figure}

\textbf{Implementation.} The pre-trained model is trained with Adam optimizer ($\beta_1=0.9, \beta_2 = 0.99$) of a minibatch of 12 for 200 epochs. The initial learning rate is set to 1e-4. We gradually decrease the learning rate via the cosine policy~\cite{2016SGDR}. 
Following existing works ~\cite{wang2019dynamic, zheng2020parameter}, we adopt position jittering as data augmentation and set $k=20$ for the KNN algorithm to build the dynamic graph. 
In the classification network, to fairly compare with other works, we deploy four EdgeConv layers and the channel number is \{64, 64, 128, 256\} by default. In the segmentation network, we adopt three EdgeConv layers and the channel number is \{64, 64, 64\}. Two dropouts with 0.5 drop rate are used when fusing the conditional coordinates and the embedding vector. We deploy one Nvidia RTX 3090 by default. The pre-training time is about one day. FLOPs (Floating Point Operations) of the whole model is 2.798 GFLOPs and the number of parameters is 2.070~\emph{M}. The test time is about 0.0019 seconds per point cloud. FLOPs and the number of parameters of the decoder are 0.729 GFLOPs and 0.712 \emph{M}. It is worth noting that the complexity of our model mainly depends on the encoder structure. Our method is open to different encoder choices, which can be selected according to the computing resource. 
The proposed method can converge smoothly as shown in Fig.~\ref{fig:loss_curve}. In the fine-tuning stage, we add a linear classifier for both classification and segmentation tasks. The segmentation task demands extra category vector (see Fig.~\ref{fig:encoder-decoder}). We keep a random category vector during pre-training to hold the position and provide the real category vector during fine-tuning.

\subsection{Quantitative \& Qualitative Results}

\begin{figure}[t]
\begin{center}
\includegraphics[width=0.95\linewidth]{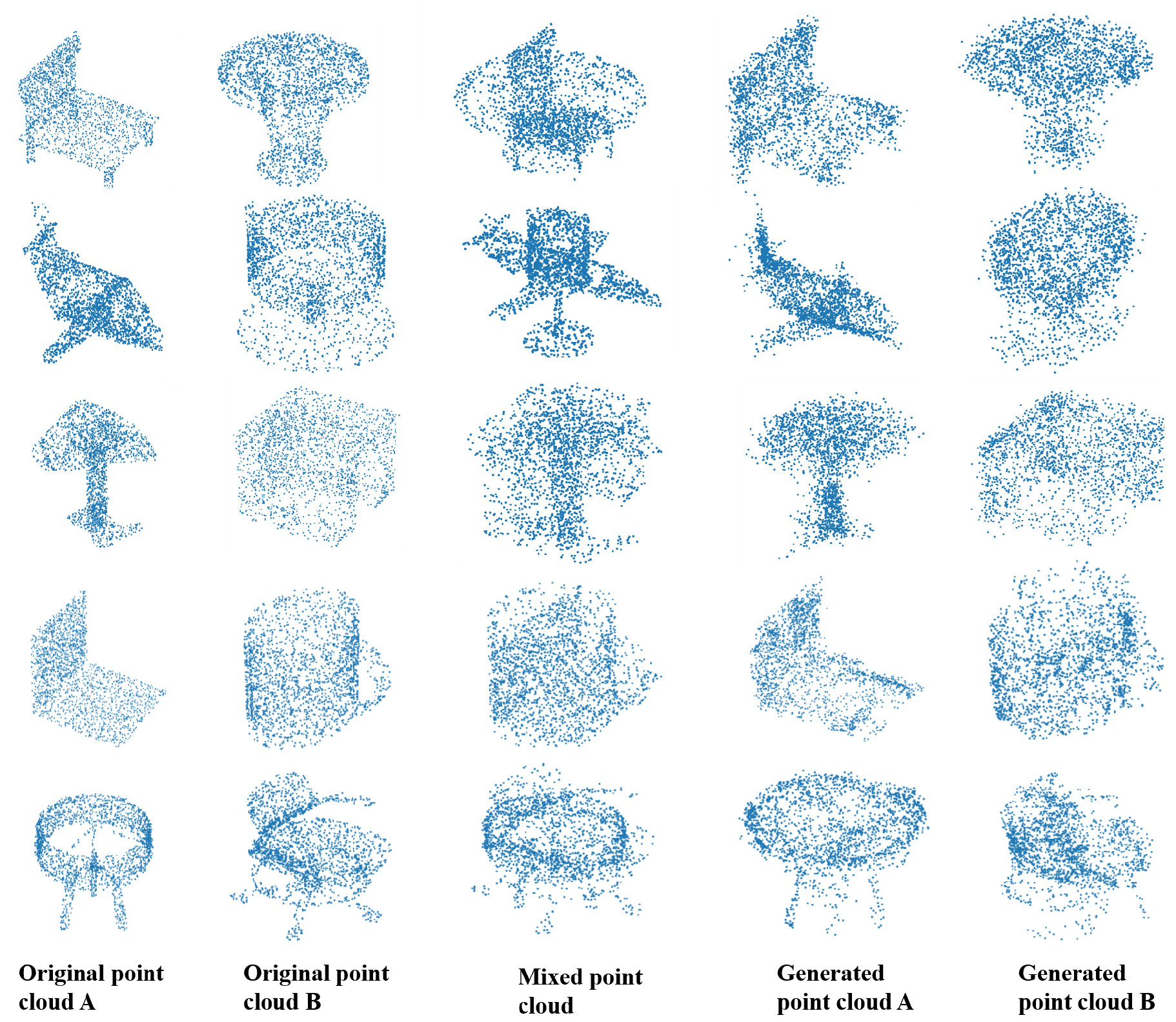}
\end{center}
\vspace{-.15in}
   \caption{Visualization of recovered point clouds after disentangling. The experiment is performed on the Shapenet-Part. 
   We observe that the proposed method can successfully disentangle the mixed point cloud into two separate point clouds. In line 5, the chair and the table have a similar structure. There is a hole in the middle of the table and the model tends to fill this hole which makes the disentangling more challenging.}
\label{fig:disentangle}
\end{figure}



\begin{figure}[t]
\begin{center}
\includegraphics[width=0.95\linewidth]{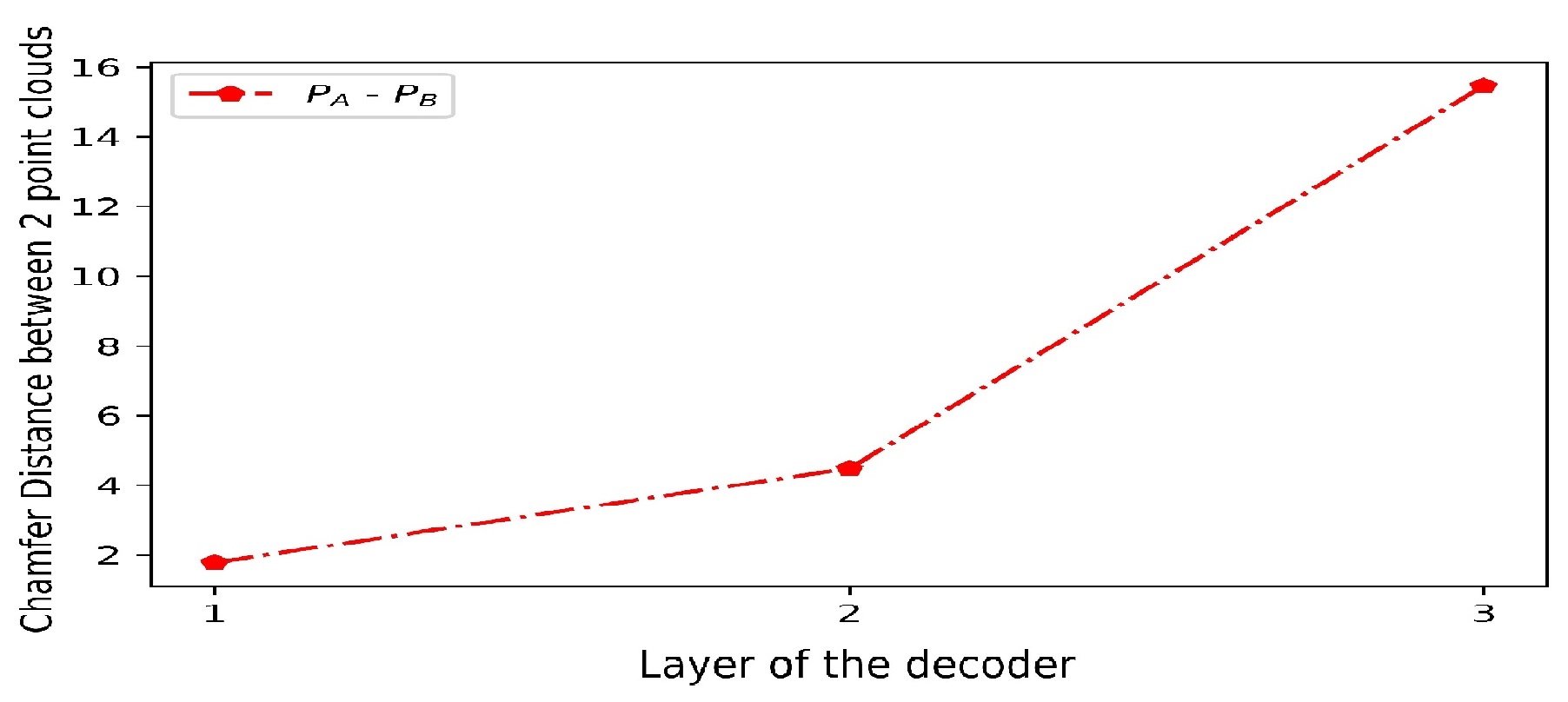}
\end{center}
\vspace{-.15in}
   \caption{Visualization of the distance between input point clouds $P_A$ and $P_B$ in 3 layers of the decoder. We observe that the distance between $P_A$ and $P_B$ becomes larger from the shallow layer to the deep layer in the decoder.}
\label{fig:feature}
\end{figure}


\begin{figure}[t]
\begin{center}
\includegraphics[width=0.9\linewidth]{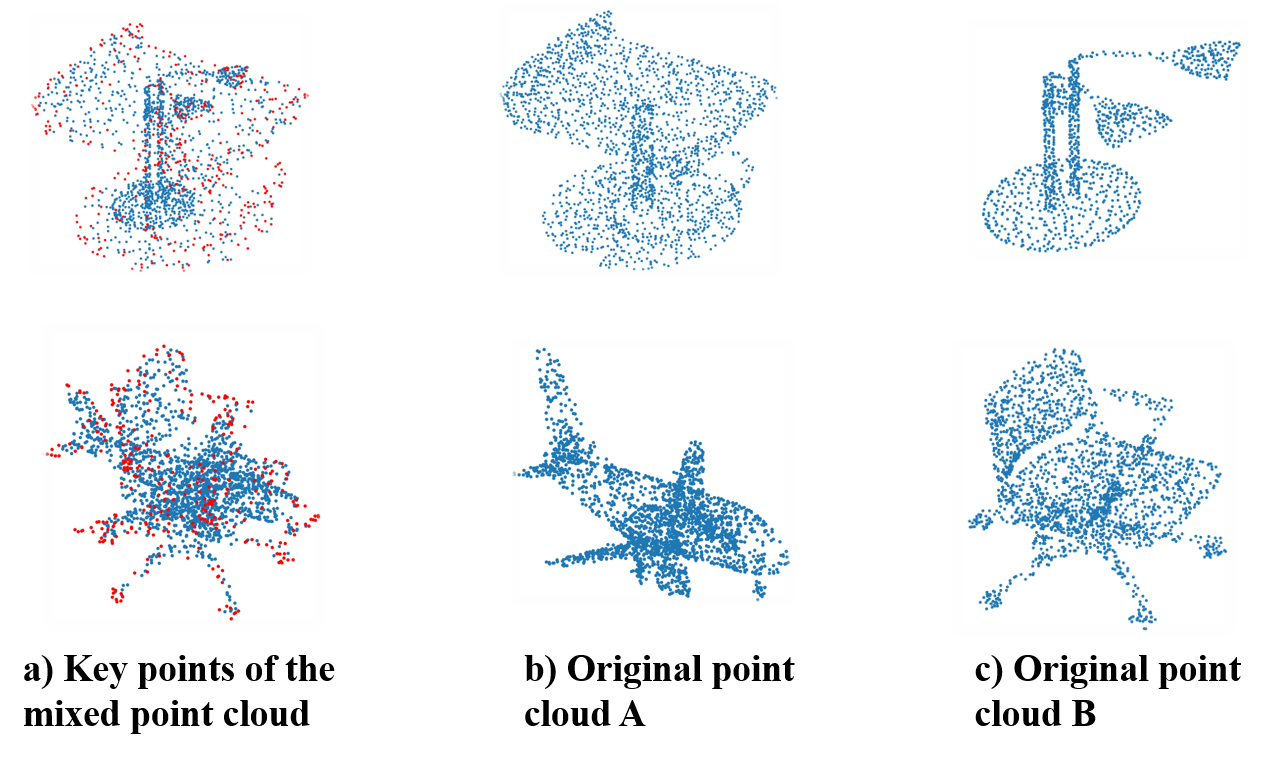}
\end{center}
\vspace{-.15in}
   \caption{Visualization of the key points of the mixed point cloud. a) is the point cloud mixed by b) and c), and the key points of a) is indicated by red points. The result shows that our encoder can successfully capture the key points of two original point clouds after pre-training.} 
\label{fig:keypoints}
\end{figure}

\textbf{The pre-training improves the baseline.} We evaluate how the pre-training affects the performance in Tab.~\ref{table:pre-train}. For the classification task, the model is pre-trained and fine-tuned on ModelNet-40. We note that the pre-trained model gets an overall accuracy of 93.39\%, which have +0.65\% than the baseline (92.74\%). Similarly, the mean class accuracy also increases from 89.88\% to 90.26\% with +0.38\% accuracy improvement. For the segmentation task, the model is pre-trained and fine-tuned on ShapeNet-Part. The pre-trained model achieves a competitive mean Intersection-over-Union~(mIoU) of 85.50\%. Experimental results show that the accuracy of our baseline surpasses three existing works, and pre-training can still bring gains to the accuracy both in classification and segmentation tasks, which verifies the effectiveness of our pretext task. 


\begin{table}[t]
\footnotesize
\begin{center}
\vspace{-0.2cm}
\setlength{\tabcolsep}{1.0mm} {
\scalebox{0.95}{
\begin{tabular}{l|c|c|c|c}
\shline
\multirow{3}{*}{Methods} & 
\multicolumn{2}{c|}{ModelNet-40} &
\multicolumn{1}{c|}{ShapeNet-Part} &
\multicolumn{1}{c}{S3DIS}
 \\
& \multicolumn{2}{c|}{\emph{Classification}} &
\multicolumn{1}{c|}{\emph{Segmentation}} &
\multicolumn{1}{c}{\emph{Segmentation}}
 \\
\cline{2-5}
& OA~(\%) & mA~(\%) & mIoU~(\%) & mIoU~(\%) \\
\shline
Ours~(from scratch) & 92.74 & 89.88 & 85.27 & 50.78  \\
Ours~$^*$ & \textbf{92.78} & \textbf{90.06} & \textbf{85.37} & \textbf{51.74}  \\
\shline
\end{tabular}
}
}
\end{center}
\caption{\textcolor{black}{The performance of our method in the real-scan dataset. We pre-train the model on S3DIS and fine-tune the model in 3 different tasks. The results suggest a consistent improvement.}} 
OA: Overall Accuracy; ~
mA: Mean Class Accuracy; ~
mIoU: mean IoU; ~
$^*$ means pre-training on S3DIS
\label{table:s3dis}
\end{table}

\setlength{\tabcolsep}{3pt}
\begin{table}[t]{
\caption{Segmentation results on partial labeled data. 
The performance boost is more significant when labeled data is limited, 
verifying our intuition to benefit the real-world model training under the data limitation. } \label{table:rarelabel}
{
\vspace{-0.2cm}
\scalebox{0.73}{
\begin{tabular}{c|c|c|c}
\shline
Labeled Data Ratio~(\%) & Pre-trained & Average Accuracy~(\%) & mIoU~(\%) \\
\shline
10  &  & 74.02 & 81.59 \\
10  & $\checkmark$ & 77.04 & 82.23 \\
\hline

25 &  & 80.39 & 83.76 \\
25 & $\checkmark$ & 81.11 & 84.49 \\
\hline
50  & & 81.46 & 84.71 \\
50  & $\checkmark$ & 83.31 & 85.04 \\
\shline
\end{tabular}}
}}
\end{table}

\setlength{\tabcolsep}{3pt}
\begin{table}[t]{
\caption{\textcolor{black}{Semi-supervised segmentation results on 10\% labeled data. We can observe two points: (1) Both pseudo label-based approach and our method can improve the performance. (2) Our method can be coupled with other semi-supervised learning methods to improve the performance.}} \label{table:semi2}
{
\vspace{-0.2cm}
\scalebox{0.90}{
\begin{tabular}{l|c|c|c}
\shline
Methods & Pre-trained & Average Accuracy (\%) & mIoU~(\%) \\
\shline
baseline  &  & 74.02 & 81.59 \\
baseline  &  $\checkmark$ & \textbf{77.04} & \textbf{82.23} \\
\hline
pseudo label &   & 72.39 & 82.50 \\
pseudo label &  $\checkmark$ & \textbf{73.57} & \textbf{82.81} \\
\shline
\end{tabular}}
}}
\end{table}

\begin{figure}[t]
\begin{center}
\includegraphics[width=0.9\linewidth]{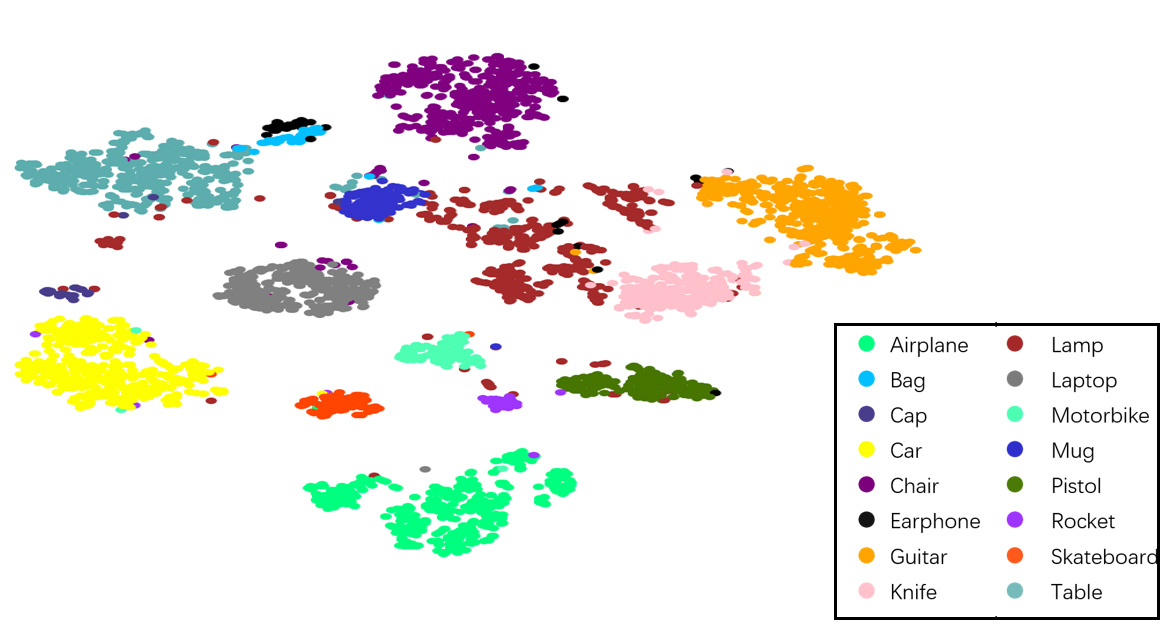}
\end{center}
\vspace{-.2in}
  \caption{The visualization of embeddings on ShapeNet-Part in the semantic space via T-SNE~\cite{van2008visualizing}. The encoder is pre-trained on ModelNet-40 and is applied to encode the point clouds on ShapeNet-Part directly. The results show that the distance between intra-class samples is small while the distance between inter-class data is large, which verifies that our encoder learns a robust and general structure embedding vector. 
   }
\label{fig:cluster}
\end{figure}

\textbf{Visualization of the reconstructed point cloud.} (1) As shown in Fig.~\ref{fig:disentangle}, we visualize original point clouds, mixed point clouds, and generated point clouds on ShapeNet-Part. 
Our method achieves good reconstruction results, recovering the key geometry. 
(2) We further track the decoder activation changes during reconstruction.  Given one mixed embedding (extracted from mixed $P_A$ and $P_B$), we visualize the disentangle process via Chamfer distance (see Fig.~\ref{fig:feature}). 
There are three layers in our decoder and the channel of each layer output is 128, 64, and 3.  We normalize the output feature in the channel dimension and divide the Chamfer distance by the channel dimension for distance calculation. We observe that the distance between $P_A$ and $P_B$ becomes larger from the shallow layer to the deep layer in the decoder. It indicates that the decoder follows the conditional information and chooses different features for reconstructing $P_A$ and $P_B$ respectively.

\textbf{Visualization of key points.} We visualize the top 25\% key points of the mixed point cloud, which has max activation after the last pooling layer in the encoder (see Fig.~\ref{fig:keypoints}). The result shows that the key points of a) include the key points of both b) and c), which verifies that the pre-trained embedding learns the salient geometric information of both input point clouds. 

\textbf{Visualization of embeddings.} 
To verify the scalability of learned embeddings, we extract embeddings of ShapeNet-Part via the pre-trained model on ModelNet-40. We apply the T-SNE~\cite{van2008visualizing} to reduce the dimension of the embedding vector to $\mathbb{R}^2$ for plotting. As shown in Fig.~\ref{fig:cluster}, the distance between intra-class samples is small and the distance between inter-class is large. The above result supports the generalization of the pre-trained encoder to unseen point clouds. 

\subsection{Ablation Studies and Further Discussion}

\textbf{Performance on the real-scan dataset.}
We further conduct our pre-training process on S3DIS, a real-scan dataset. S3DIS has 6 areas and we follow existing works~\cite{zhao2021point} using area 5 for testing and others for training. We pre-train our model in the training part and then we fine-tune the pre-trained model on ModelNet-40 for the classification task, on ShapeNet-part for the part segmentation task and on S3DIS for the segmentation task. As shown in Tab.~\ref{table:s3dis}, our method benefits the performance both on ModelNet-40 and ShapeNet-part. It verifies the scalability of the model pre-trained on the large dataset. 
Furthermore, in the real-scan dataset, our model with pre-training achieves better performance 51.74\% mIoU than the model trained from scratch (50.78\%) by a clear margin.

\begin{figure}[t]
\begin{center}
\includegraphics[width=0.95\linewidth]{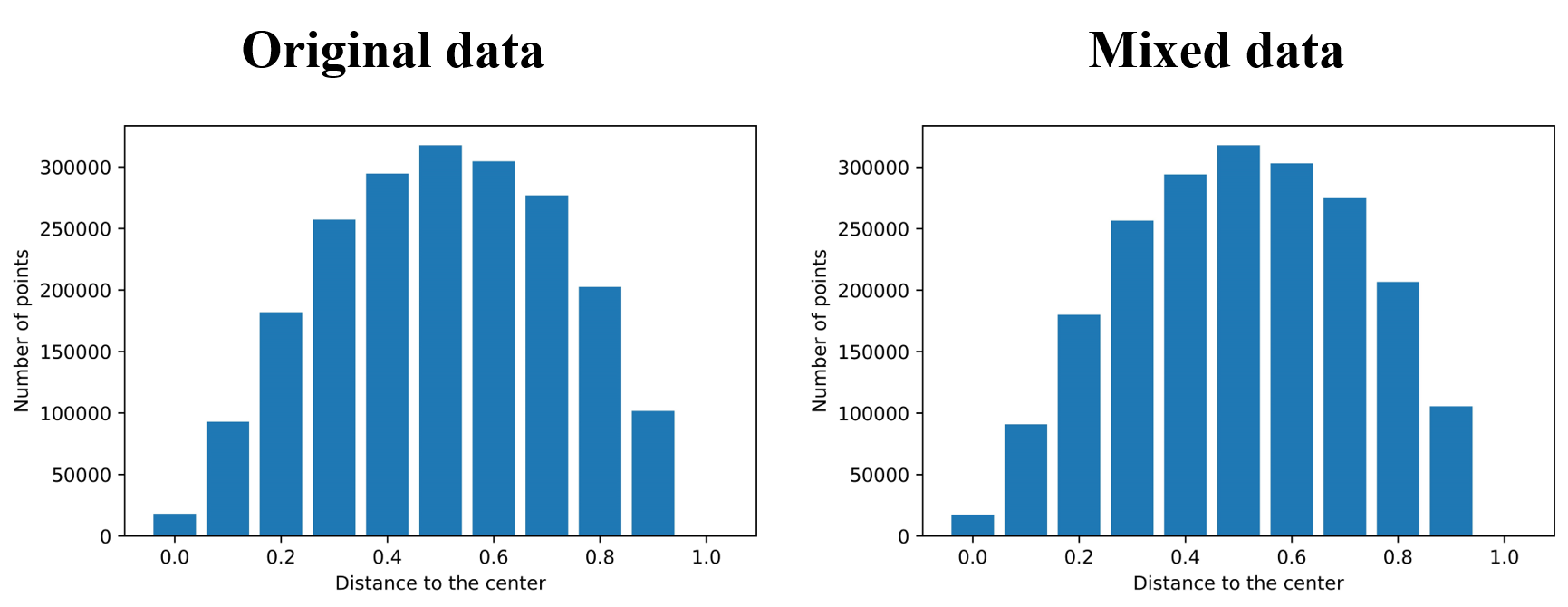}
\end{center}
\vspace{-.15in}
   \caption{\textcolor{black}{We sample 1000 point clouds and count the distance from each point to the center of its point cloud. The X-axel means the distance from each point to the center point and the Y-axel means the number of points belonging to each distance interval. We observe that both the mean and std by mixing 3D point clouds are identical to the original data distribution.}
}
\label{fig:mixdata}
\end{figure}

\textbf{Illustration of {``within-distribution’’} data.}
The “within-distribution” means that generated point clouds have similar characteristic (\eg, mean and variance) with the original point clouds. Some existing works apply GAN to generate 3D point clouds. This line of works may change the data distribution (mean and variance), since 3D point clouds are typically generated from 
random Gaussian noise~\cite{zhang2021unsupervised, shu20193d, li2019pu}. Compared with these GAN-based methods, new point clouds generated by our method are more similar to the original ones from a statistics view. We sample 1000 point clouds from ModelNet-40 and plot the distance between each point to its center point. 
In Fig.~\ref{fig:mixdata}, we observe that the distribution between the original data and mixed 3D point clouds is almost identical.


\setlength{\tabcolsep}{15pt}
\begin{table}[t]{
\caption{
We select 20 classes of 40 categories on ModelNet-40 for pre-training, and the rest 20 classes for fine-tuning and testing.
The accuracy of our method surpasses the model trained from scratch by a clear margin.} \label{table:rest}
{
\scalebox{0.9}{
\begin{tabular}{l|c|c}
\shline
Methods & OA (\%) & mA (\%) \\
\shline
baseline  & 95.07 & 93.82 \\
\hline
pre-training  &  \textbf{96.65} & \textbf{94.14} \\
\shline
\end{tabular}}
}}
\end{table}

\textbf{Comparisons under the semi-supervised setting.} In some real scenarios, labeled data is inadequate. To simulate this harsh situation, we divide the original dataset into two parts $A$ and $B$, and regard $A$ as unlabeled data. 
We use $A+B$ to pre-train the model and use $B$ to fine-tune the model.
\textbf{(1)} We set the percentages of labeled data as 10\%, 25\%, and 50\% respectively. We verify the effect of pre-training by comparing the encoder trained from scratch with the encoder pre-trained on ShapeNet-Part in Tab.~\ref{table:rarelabel}. 
The results show that as the amount of labeled data increases, the accuracy of segmentation gradually increases. The performance of the pre-trained model generally surpasses that of the model trained from scratch, especially when the labeled data is extremely limited. This verifies that our method can successfully leverage unlabeled data to improve the accuracy of the model with limited annotated data. 
\textbf{(2)} Although our work is not designed for the semi-supervised setting, it can nevertheless be coupled with other semi-supervised learning methods. 
To verify this point, we build a preliminary semi-supervised baseline based on predicted pseudo labels. We adopt 10\% labeled data to train the baseline model. Pseudo labels for the rest 90\% unlabeled data are then predicted by this baseline model.  
We select the unlabeled data with the pseudo label confidence greater than 0.7 and the original 10\% labeled data to form the new training set, and then fine-tune the segmentation model. 
In Tab.~\ref{table:semi2}, the model trained on the pseudo-labeled data improves the mIoU performance from 81.59\% to 82.50\%, but is suffer from the label noise, which compromises average accuracy. Our pre-training method can be coupled with the pseudo label to relieve the negative impact from noisy annotations. 
Specifically, initializing the model with our pre-training weight can help to predict more robust pseudo labels. Therefore, semi-supervised methods with our pre-training can arrive at better average accuracy from 72.39\% to 73.57\% and mIoU from 82.50\% to 82.81\%.
\textbf{(3)} We further study category-wise semi-supervised setting. In particular, we select 20 classes of 40 categories on ModelNet-40 for pre-training, and the rest 20 classes for fine-tuning and testing, while the compared baseline model is training and testing only on the remaining 20 classes. We compare the performance of the model initialized randomly (baseline) and initialized with our pre-training method in Tab.~\ref{table:rest}. The accuracy of our method surpasses the model trained from scratch by a clear margin.

\textbf{Scalability of learned embeddings.} We apply a simple linear classifier to directly classify learned embeddings. The classifier consists of a linear layer, a batch norm layer and a linear layer. We arrive at a competitive accuracy of 89.63\% with the fine-tuning result of 93.39\% on ModelNet-40. Similar to the observation in the large pre-training model CLIP~\cite{radford2021learning}, fine-tuning helps the pre-trained model to achieve better performance on specific sub-tasks, while directly using learned features also is a sub-optimal but efficient choice.


\textbf{Effect of the learnable aggregation.} We deploy LA to aggregate neighbor point features. We apply the same network architecture to compare the accuracy of LA, max pooling, and attentive pooling~\cite{hu2020randla}. As shown in Tab.~\ref{table:Contrastive loss}, regardless of pre-training, the accuracy of using LA is higher than that of using max pooling or attentive pooling. 
The observation verifies that LA can better preserve the local context of the point cloud. There are 3 LA modules in our model and concrete numbers of learned $\alpha$ in each layer are  0.6996, 0.6503, and 0.5500 from shallow layer to deep layer in the encoder. It means that in shadow layers, max pooling plays an important role to reduce the redundant points, and in deep layers, average pooling is as important as max pooling to aggregate the feature.

\textbf{Effect of the contrastive loss.}
We apply the contrastive loss to narrow the distance between similar samples and widen the distance between different samples. The encoder can extract more robust embedding from data, which can improve downstream tasks. We train two pre-trained models with or without the contrastive loss and then fine-tune the two models on ShapeNet-Part, and finally, compare the segmentation accuracy of the two models in Tab.~\ref{table:Contrastive loss}. We observe that the contrastive loss can further improve the segmentation performance. 

\begin{figure}[t]
\begin{center}
\includegraphics[width=0.95\linewidth]{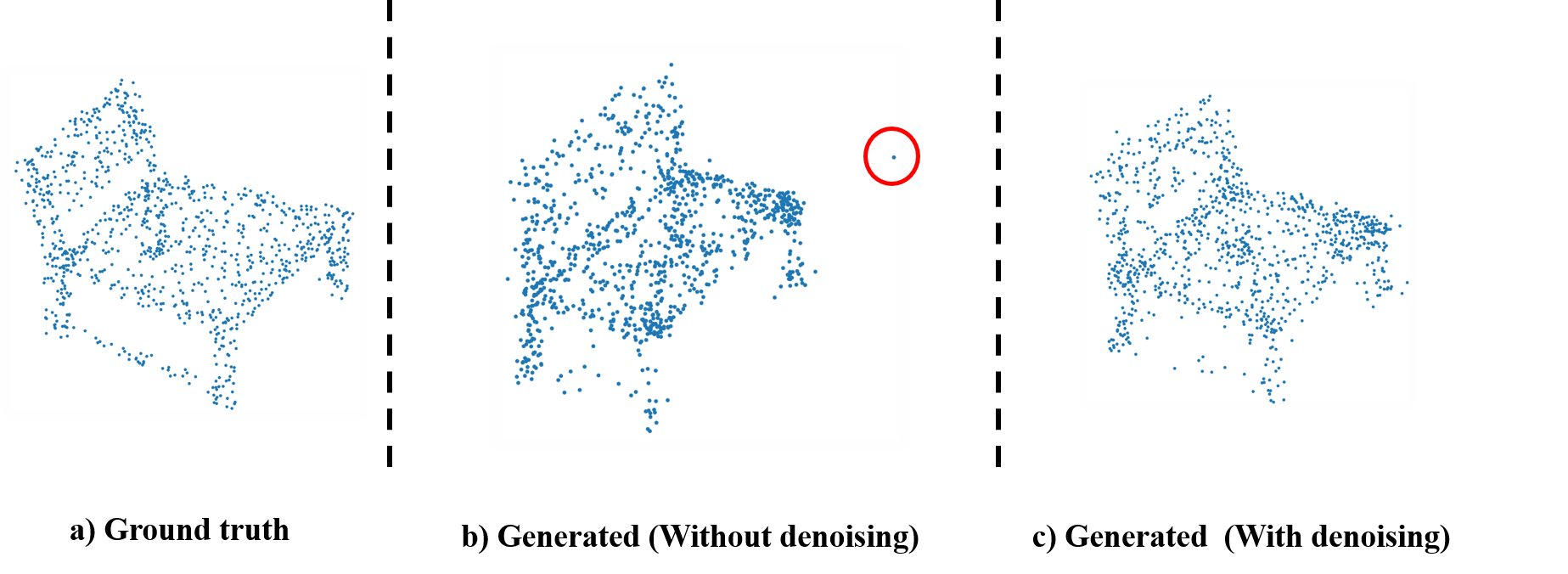}
\end{center}
\vspace{-.15in}
   \caption{Visualization results of generated point clouds. The original point cloud is shown in a). b) is the generated point cloud without denoising and c) is the point cloud after denoising. The result shows that the denoise module can help to remove the noise point circled in b).}
\label{fig:denoise}
\end{figure}

\begin{table}[tbp]
\footnotesize
\caption{Ablation study on different components during training. We verify the effects of LA module and Contrastive loss. 
}   
\label{table:Contrastive loss}
\begin{center}
\resizebox{\linewidth}{!}
{
\setlength{\tabcolsep}{3pt}
\begin{tabular}{c|c|c|c}
\shline
Pooling Method & Pre-trained & Contrastive loss & mIoU~(\%) \\
\shline
Max Pooling & & & 85.17 \\
Max Pooling & $\checkmark$ & & 85.34 \\
Attentive Pooling~\cite{hu2020randla} & & & 85.12 \\ 
{\color{black}Attentive Pooling} & {\color{black}$\checkmark$} & {\color{black}$\checkmark$} & {\color{black}85.36} \\ 
\hline
LA & & & 85.27 \\
LA & $\checkmark$ & & 85.40 \\
LA & $\checkmark$ & $\checkmark$ & 85.50 \\ 

\shline
\end{tabular}}
\end{center}
\end{table}

\begin{table}[t]
\footnotesize
\setlength{\tabcolsep}{2.8pt}
\begin{center}
\vspace{-0.2cm}
\setlength{\tabcolsep}{2.3mm} {
\begin{tabular}{l|c|c|c}
\shline
\multirow{3}{*}{Methods}&
\multirow{3}{*}{Pre-trained}& 
\multicolumn{2}{c}{ModelNet-40} \\
& & \multicolumn{2}{c}{\emph{Classification}} \\
\cline{3-4}
& & OA~(\%) & mA~(\%) \\
\shline
Ours & -             & 92.74 & 89.88 \\
Ours & ShapeNet-Part & 92.79 & 90.10 \\
Ours & ModelNet-40   &  \textbf{93.39} & \textbf{90.26} \\
\hline
PointNet++ ~\cite{qi2017pointnet++} $^*$ & - & 92.07 & 88.89 \\
PointNet++  $^*$ + Ours & ShapeNet-Part & 92.19 & 89.63 \\
PointNet++  $^*$ + Ours & ModelNet-40 & \textbf{92.57} & \textbf{89.96} \\
\hline
OGNet$^*$ ~\cite{zheng2020parameter} & - & 93.23 & 89.82 \\
OGNet$^*$ + Ours  & ShapeNet-Part & \textbf{93.35} & 90.51 \\
OGNet$^*$ + Ours & ModelNet-40 & 93.31 & \textbf{90.71} \\
\hline
{\color{black}PAConv$^*$~(PN)} ~\cite{xu2021paconv} & {\color{black}-} & {\color{black}92.50} & {\color{black}-} \\
{\color{black}PAConv$^*$~(PN) + Ours}  & {\color{black}ShapeNet-Part} & {\color{black}92.70} & {\color{black}-} \\
{\color{black}PAConv$^*$~(PN) + Ours} & {\color{black}ModelNet-40} & {\color{black}\textbf{92.79}} & {\color{black}-} \\

\hline
{\color{black}PT$^*$} ~\cite{zhao2021point} & {\color{black}-} & {\color{black}91.47} & {\color{black}89.32} \\
{\color{black}PT$^*$} + {\color{black}Ours}  & {\color{black}ShapeNet-Part} & {\color{black}92.03} & {\color{black}89.50} \\
{\color{black}PT$^*$ + Ours} & {\color{black}ModelNet-40} & {\color{black}\textbf{92.07}} & {\color{black}\textbf{89.58}} \\

\hline
{\color{black}PCT$^*$} ~\cite{guo2021pct} & {\color{black}-} & {\color{black}92.71} & {\color{black}89.36} \\
{\color{black}PCT$^*$ + Ours}  & {\color{black}ShapeNet-Part} & {\color{black}93.07} & {\color{black}90.27} \\
{\color{black}PCT$^*$ + Ours} & {\color{black}ModelNet-40} & {\color{black}\textbf{93.15}} & {\color{black}\textbf{90.56}} \\
\shline
\end{tabular}}
\end{center}
\caption{
Results of using 
different backbones.
We can obtain one consistent result that our method is compatible with different network structures and can still improve the accuracy of classification and segmentation tasks.
} 
\footnotesize{
PT donates Point Transformer. PCT donates Point Cloud Transformer. PAConv~(PN) donates using PointNet as the backbone~(without voting).
OA: Overall Accuracy;
mA: Mean Class Accuracy;
mIoU: mean IoU;
$^*$: We re-implement the model,
which achieves a slightly different performance.
}
\label{table:pointnet2}
\end{table}

\textbf{Effect of DenoiseBlock.} The visualization results are shown in  Fig.~\ref{fig:denoise}. a) is the original point cloud, and b) is the generated point cloud obtained by the network without the denoising module. c) is the generated point cloud obtained by the network with the denoising module. We can find that there is a noise point circled in the red circle of b). 
The denoising module successfully pulls the noise point back to the point cloud by considering the global feature. 

\textbf{Compatibility with different backbone structures.} To verify that our pretext task is free from the choice of backbones, we further explore leveraging several widely-adopted architecture as the backbone of the point cloud encoder, such as PointNet++~\cite{qi2017pointnet++}, OGNet~\cite{zheng2020parameter}, PAConv~\cite{xu2021paconv}, Point Transformer~\cite{zhao2021point} and Point Cloud Transformer~\cite{guo2021pct}.
We compare the accuracy of the model trained from scratch and initialized with the pre-trained model. We carry out experiments of classification respectively and results are shown in Tab.~\ref{table:pointnet2}. We observe consistent improvements with these backbones which verifies the generality of the proposed method. 


\section{Conclusion}
\label{sec:conclusion}
In this paper, we propose a new self-supervised learning method, called Mixing and Disentangling~(MD), for point cloud pre-training. Different from existing works, we propose to mix the original point clouds in the training set to form ``new'' data and then demand the model to ``separate'' the mixed point cloud. In this way, the model is asked to mine the geometric knowledge, \eg, the shape-related key points for reconstruction. To verify the effectiveness of the proposed method, we build a simple baseline to implement our method. We use an encoder to obtain the embedding of the mixed point cloud and then an instance-adaptive decoder is harnessed to separate the original point clouds. During the self-supervised training, the encoder learns the prior knowledge of point cloud structure, which is scalable and can improve the downstream tasks. Experiments show consistent performance improvement through classification and segmentation tasks and verify the effectiveness of our method especially when the number of labeled data is limited. 
We hope that our approach can benefit the future point cloud works, and take one closer step to the harsh real-world setting, \ie, limited annotations. In the future, we will continue to study the point cloud pre-training method on large-scale datasets, and focus on finding an efficient way to take advantage of the large-scale point data.

\bibliographystyle{IEEEtran}
\bibliography{egbib}
\begin{IEEEbiography}[{\includegraphics[width=1in,height=1.2in,clip,keepaspectratio]{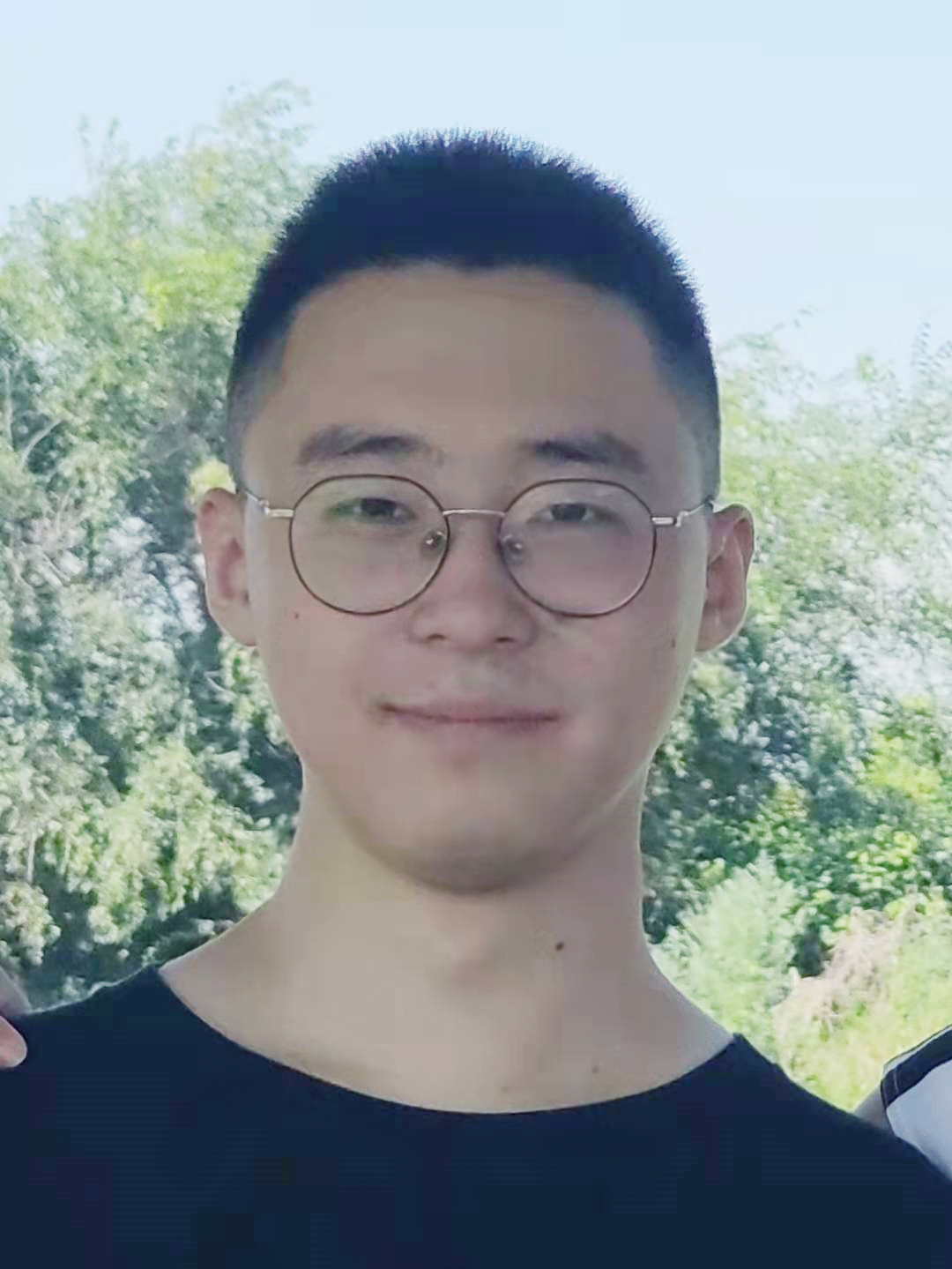}}]{Chao Sun} received the B.E. degree in software engineering from Zhejiang University, China, in 2021. He is currently a master student with the School of Computer Science at Zhejiang University, China. His current research interests include the 3D point cloud and the generative models. 
\end{IEEEbiography}

\begin{IEEEbiography}[{\includegraphics[width=1in,height=1.2in,clip,keepaspectratio]{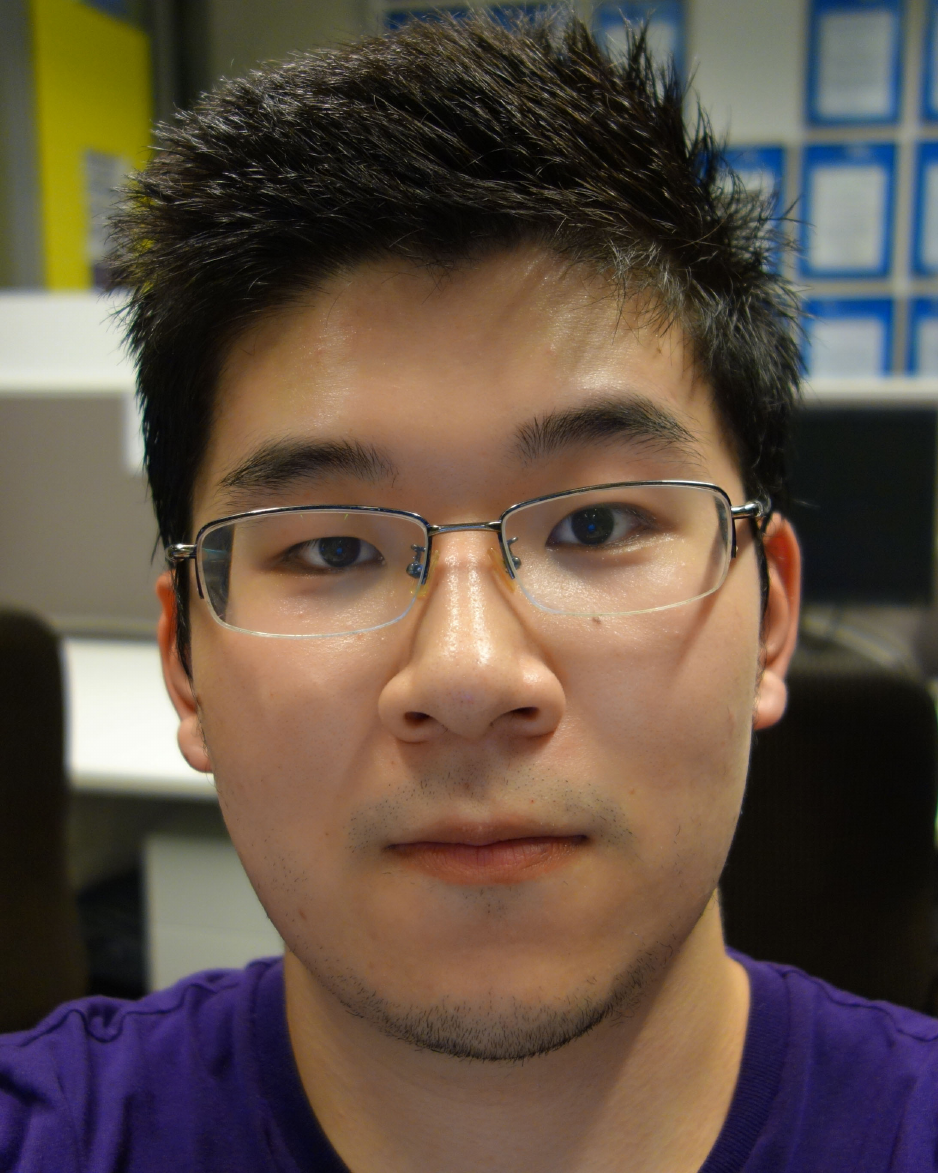}}]{Zhedong Zheng} received the Ph.D. degree from the University of Technology Sydney, Australia, in 2021 and the B.S. degree from Fudan University, China, in 2016. He is currently a postdoctoral research fellow at Sea-NExT joint lab, School of Computing, National University of Singapore. He was an intern at Nvidia Research (2018) and Baidu Research (2020). His research interests include robust learning for image retrieval, generative learning for data augmentation, and unsupervised domain adaptation.
\end{IEEEbiography}

\begin{IEEEbiography}[{\includegraphics[width=1in,height=1.2in,clip,keepaspectratio]{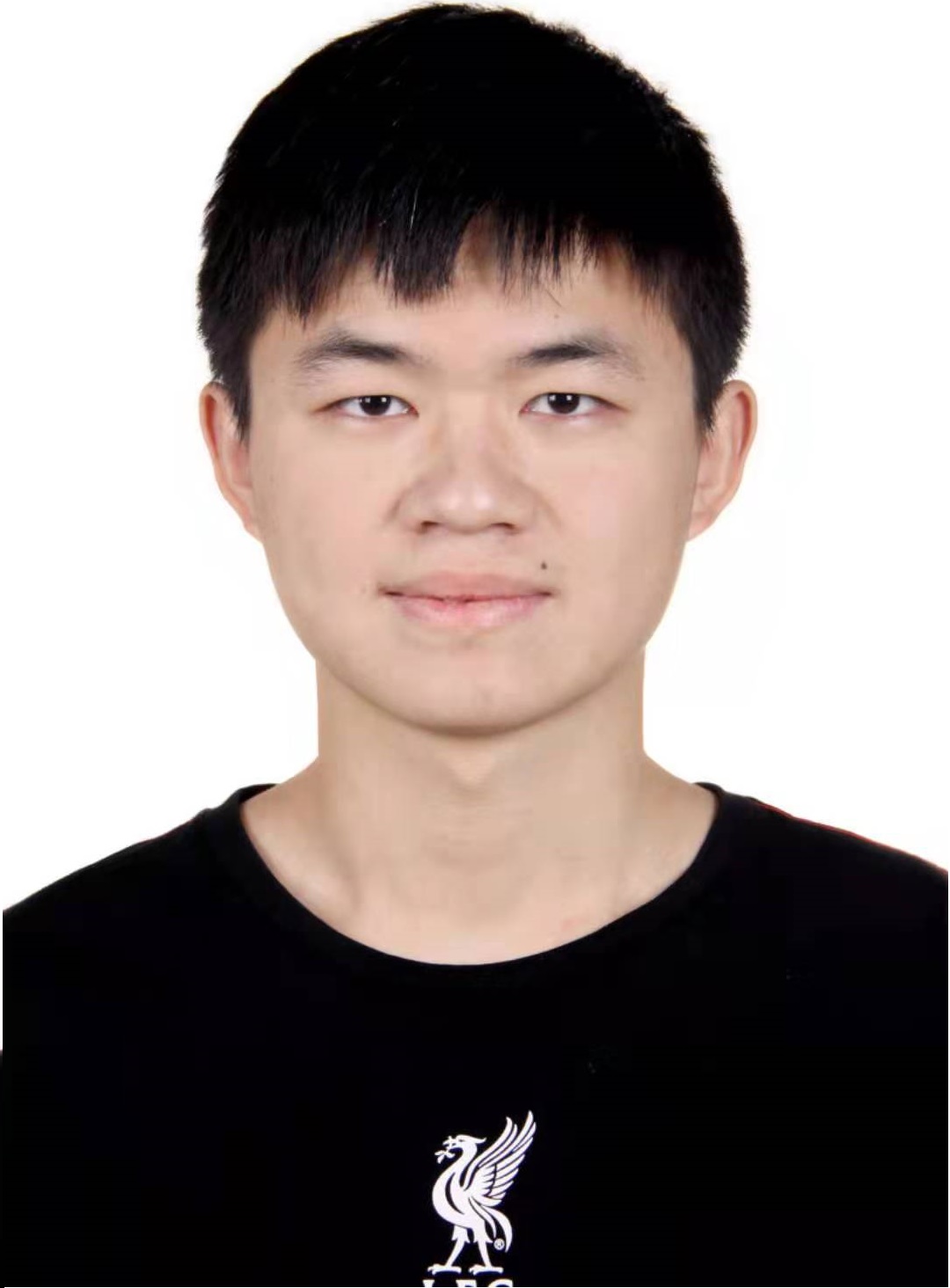}}]{Xiaohan Wang}
received the Ph.D. degree in computer science from University of Technology Sydney, Australia, in 2021. He received the B.E. degree from University of Science and Technology of China, China, in 2017. He is currently a postdoctoral researcher with the College of Computer Science and Technology, Zhejiang University, China. His research interest includes video analysis, egocentric vision and multi-modal understanding.
\end{IEEEbiography}

\begin{IEEEbiography}[{\includegraphics[width=1in,height=1.2in,clip,keepaspectratio]{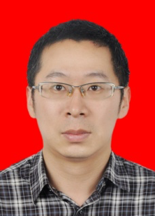}}]{Mingliang Xu}  is a professor in the School of Information Engineering of Zhengzhou University, China. He received his Ph.D. degree in computer science and technology from the State Key Lab of CAD\&CG at Zhejiang University, Hangzhou, China, and the B.S. and M.S. degrees from the Computer Science Department, Zhengzhou University, Zhengzhou, China, respectively. 
\end{IEEEbiography}

\begin{IEEEbiography}[{\includegraphics[width=1in,height=1.2in,clip,keepaspectratio]{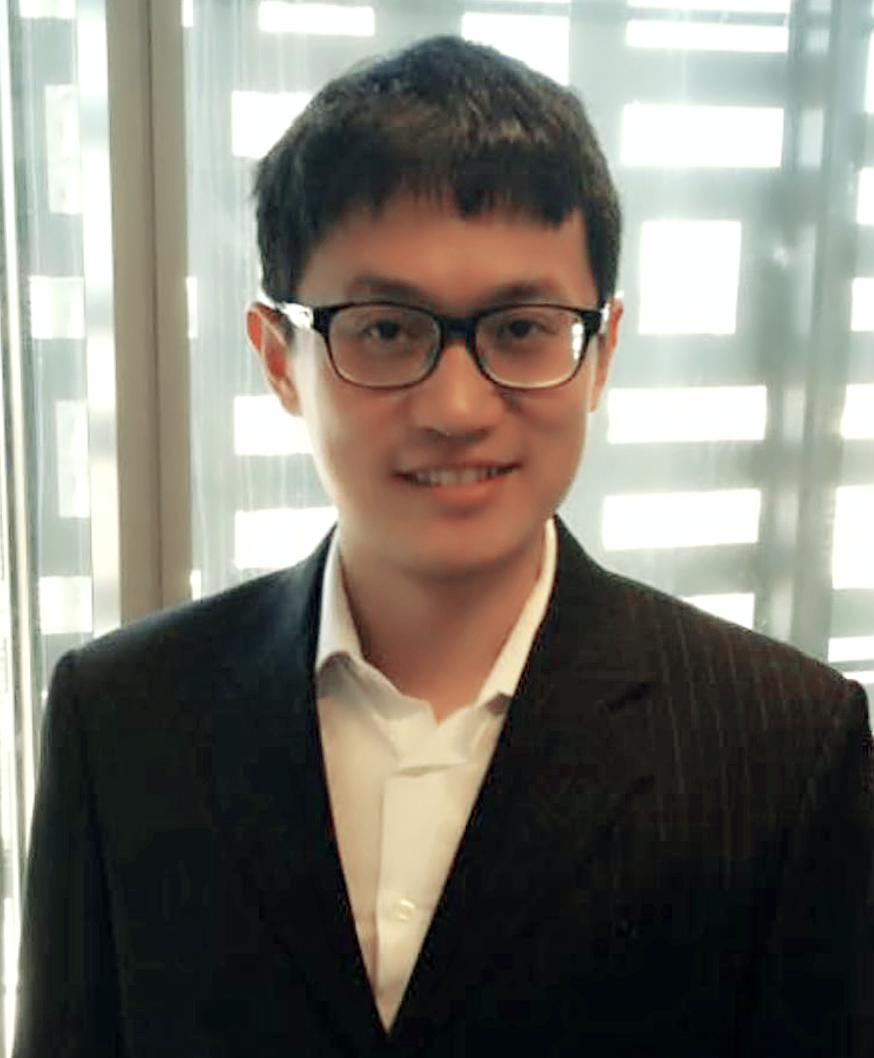}}]{Yi Yang} received the Ph.D. degree in computer
science from Zhejiang University, Hangzhou, China, in 2010. He is currently a professor with University of Technology Sydney, Australia.
He was a Post-Doctoral Research with the School of Computer Science, Carnegie Mellon University, Pittsburgh, PA, USA. His current research interest includes machine learning and its applications to multimedia content analysis and computer vision. 
\end{IEEEbiography}

\end{document}